\documentclass[11pt]{article}

\usepackage{acl}

\usepackage{times}
\usepackage{latexsym}
\usepackage[T1]{fontenc}
\usepackage[utf8]{inputenc}
\usepackage{microtype}
\usepackage{inconsolata}
\usepackage{url}
\usepackage{booktabs}
\usepackage{amsfonts}
\usepackage{amsmath}
\usepackage{amssymb}
\usepackage{amsthm}
\usepackage{nicefrac}
\usepackage{graphicx}
\usepackage{xcolor}
\usepackage{multirow}
\usepackage{subcaption}
\usepackage{algorithm}
\usepackage{algpseudocode}
\usepackage{placeins}
\usepackage{float}
\usepackage{tabularx}
\usepackage{makecell}
\usepackage{threeparttable}
\usepackage{siunitx}
\usepackage{array}

\newcolumntype{Y}{>{\raggedright\arraybackslash}X}
\newcolumntype{L}[1]{>{\raggedright\arraybackslash}p{#1}}

\usepackage[most]{tcolorbox}
\usepackage{listings}
\usepackage{xcolor}

\lstdefinestyle{promptstyle}{
  basicstyle=\ttfamily\footnotesize,
  breaklines=true,
  breakatwhitespace=false,
  columns=fullflexible,
  keepspaces=true,
  showstringspaces=false,
  upquote=true,
  tabsize=2
}

\newtcblisting{promptbox}[1]{
  enhanced,
  breakable,
  colback=gray!4,
  colframe=gray!55,
  boxrule=0.5pt,
  arc=1.5mm,
  left=1mm,
  right=1mm,
  top=1mm,
  bottom=1mm,
  title={#1},
  fonttitle=\bfseries\small,
  listing only,
  listing style=promptstyle
}
\definecolor{AppAccent}{RGB}{38,82,132}
\definecolor{AppRule}{RGB}{205,214,226}
\definecolor{AppSoft}{RGB}{247,249,252}


\graphicspath{{figures/}{./}{./figures/}}

\newcommand{\pc}{p_{\mathrm{c}}}

\theoremstyle{plain}
\newtheorem{theorem}{Theorem}[section]

\newtheorem{corollary}[theorem]{Corollary}

\title{When Tools Hurt LLM Reasoning: State-Dependent Belief Revision under External Evidence}

\author{
Yufeng Zhao\textsuperscript{1,2,3}\thanks{Equal contribution.}
\quad
Junnan Liu\textsuperscript{4}\footnotemark[1]
\quad
Hongwei Liu\textsuperscript{1}
\quad
Dongsheng Zhu\textsuperscript{1}
\\
\textbf{Yuan Shen}\textsuperscript{2,3}\thanks{Corresponding authors.}
\quad
\textbf{Songyang Zhang}\textsuperscript{5}\footnotemark[2]
\quad
\textbf{Kai Chen}\textsuperscript{1}\footnotemark[2]
\\[2mm]
\textsuperscript{1}Shanghai Artificial Intelligence Laboratory
\\
\textsuperscript{2}Department of Electronic Engineering, Tsinghua University
\\\textsuperscript{3}Beijing National Research Center for Information Science and Technology
\\
\textsuperscript{4}Department of Data Science and AI, Monash University \quad
\textsuperscript{5}Tencent Hunyuan
\\
\texttt{zhao-yf21@mails.tsinghua.edu.cn; shenyuan\_ee@tsinghua.edu.cn}
}

\begin{document}

\maketitle
\begin{abstract}
Tool use is often assumed to monotonically improve reasoning, where external evidence is expected to help when relevant and be ignored when irrelevant. We show that this assumption fails in a state-dependent way. Across benchmarks with Python and Wikipedia tools, external evidence reliably helps when initial beliefs are weak, but can flip already-correct answers when those beliefs are strong. We frame this as a misallocation of revision authority, arguing that deferring to external evidence is suboptimal when internal support for the correct answer surpasses the tool's expected output quality. This predicts that harm should concentrate on high-confidence no-tool cases. We test this prediction with threshold localization, wrong-trace audits, and a same-clue intervention showing that revision framing changes the damage caused by misleading evidence. These findings suggest that mixed no-tool/tool-assisted inference should arbitrate authority rather than privilege tool evidence by default. As a minimal demonstration, we introduce \textsc{CASE}, a label-free controller that selects between no-tool and tool-assisted trajectories using answer-state certainty and improves over existing confidence-based arbitration baselines.The code for our experiments is available at \url{https://github.com/epsilondylan/State-Dependent-Belief-Revision}.

\end{abstract}

\section{Introduction}

Tool-integrated reasoning has become a central paradigm for extending the problem-solving capabilities of language models, with code execution, retrieval, and search increasingly incorporated as standard reasoning components \citep{feng2025retool, jin2025searchr1, lin2025understanding}. This success has reinforced a monotonic assumption about tool utility, namely that relevant evidence boosts reasoning performance whereas unhelpful noise is safely filtered out. However, a growing body of evidence shows that this view is incomplete. Models
may overuse tools even when parametric reasoning is sufficient
\citep{mallen2023when, qian2025smart}; they can be distracted by irrelevant
context or fail to robustly use retrieved information \citep{shi2023large,
liu2024lost}; and they can mis-arbitrate between internal knowledge and
conflicting external evidence \citep{wu2024clasheval, su2024conflictbank}.

We focus on a different question: how much \emph{revision authority} should external evidence receive, conditional on the model's pre-tool belief state? We show that tool use can impair reasoning even when the tool returns a correct local result. In these cases, the model over-authorizes the external signal and revises away from an already-correct, high-confidence answer. This failure  concentrates in strong-prior cases, where the model should preserve its internal belief rather than revise toward a tool-framed interpretation.

A representative stored trace illustrates this failure. On a SuperGPQA item asking for $\binom{85}{82}$, the no-tool state assigns $p_c=0.99$ to the correct combined option, recognizing that both $98770$ and $C(85,3)$ are valid. Python correctly returns \texttt{C(85,82)=98770}, \texttt{C(85,3)=98770}, and equality \texttt{True}. Nevertheless, the tool-assisted trajectory revises to the numeric-only option D, arguing that the single numerical value is more direct; its final probe assigns probability $0.99$ to this wrong option. The computation is correct and useful, but the answer is damaged by over-authorizing a revised interpretation of the answer space.

\begin{figure*}[!t]
    \centering
    \includegraphics[width=0.95\textwidth]{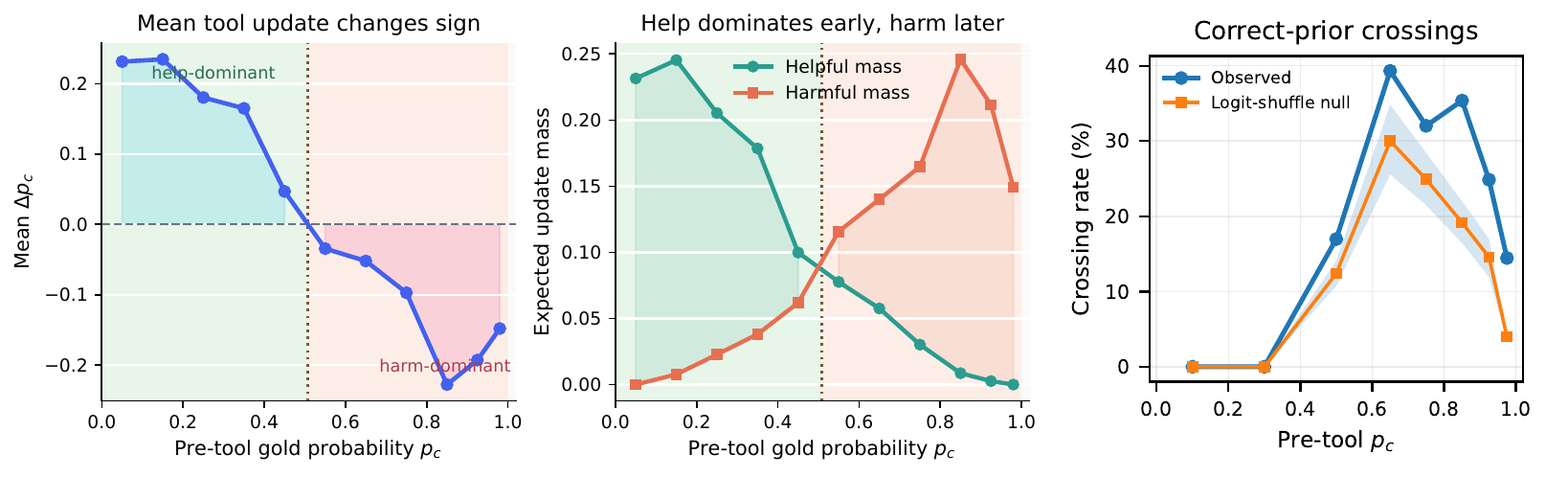}
    \caption{\textbf{State-dependent tool revision across four reasoning benchmarks.} Tool use helps weak priors but harms strong priors: mean updates change
sign with $\pc$ (left), helpful and harmful mass reverse dominance
(middle), and correct-prior crossings exceed a logit-shuffle null
(right). Additional per-dataset and log-odds diagnostics appear in
Appendices~\ref{sec:abd} and~\ref{sec:proxy_robustness}.}
    \label{fig:acs_scaling}
\end{figure*}

Across four challenging reasoning benchmarks, Figure~\ref{fig:acs_scaling} reveals a systematic prior-dependent revision profile. While tool access benefits instances with weak pre-tool confidence in the correct answer, this advantage inverts when initial beliefs are strong, manifesting as negative average updates, harmful update mass approaching helpful mass, and correct-prior crossings that exceed a geometry-preserving logit-shuffle baseline. Tool use is therefore better understood as a belief-revision intervention whose value depends on the model’s current belief state, rather than as a uniformly beneficial add-on.

We study this phenomenon as \emph{state-dependent belief revision}. The central failure is excessive revision authority assigned to external evidence. Under a minimal-change revision objective, strong priors should be protected unless the external channel provides greater expected support. This yields a threshold principle: when internal log-odds already dominate the expected support from the tool channel, the risk-minimizing external authority is zero.

We test this principle using paired no-tool and tool-assisted reasoning traces. The evidence proceeds in three steps. First, harmful flips concentrate in the high-prior regime, where internal-belief proxies exceed contextual-support proxies. Second, an error-case audit reveals why this failure easily evades surface inspection, as many failed tool-assisted traces remain deceptively supportive and highly confident. Third, the Prior-Authority Interaction Test (PAIT), a same-clue, different-frame intervention, shows that revision posture modulates the damage, while an antidote frame substantially attenuates it.

This mechanism has direct implications for deployment. When no-tool and tool-assisted trajectories are combined, equal voting can overweight the unreliable channel because it counts trajectories rather than comparing post-revision authority. We therefore formulate mixed inference as an authority-allocation problem and instantiate this view with Conflict-Aware Asymmetric Expansion (\textsc{CASE}), a label-free controller that uses answer-state certainty to arbitrate between revised and unrevised channels. \textsc{CASE} consistently improves over naive mixing in both the one-of-each setting and higher-budget expansion. Our contributions are as follows:
\begin{itemize}
\item We identify and formalize a \textbf{state-dependent belief-revision failure}: tool use can repair weak initial beliefs but may over-revise already-correct, high-confidence beliefs. The resulting threshold principle characterizes when external evidence should receive zero revision authority.
\item We support this mechanism with \textbf{three linked tests}: cross-model threshold localization of harmful flips, wrong-trace audits revealing confident and superficially supportive failures, and PAIT, a same-clue intervention showing that revision framing modulates damage.
\item We instantiate the deployment implication with \textbf{\textsc{CASE}}, a label-free controller for mixed no-tool/tool-assisted inference that allocates authority between revised and unrevised trajectories and improves over naive mixing under low-budget settings.
\end{itemize}

\section{A Threshold Law for Revision Authority}
\label{sec:theory}


The prior-dependent reversal in Figure~\ref{fig:acs_scaling} changes the
question from whether tools help on average to when external evidence should
be allowed to revise an existing belief. We formalize this question with a
minimal belief-revision model that assigns authority to the tool-conditioned
signal relative to the model's pre-tool belief. The model yields a simple
protection principle: strong internal beliefs should be preserved unless the
external channel provides stronger expected support, and turns that principle into testable predictions. Full derivations and the iterated-destruction lemma used in the wrong-trace audit appear in Appendix~\ref{app:proofs}.

\subsection{Setup and Log-Odds Identity}

Given a question $x$ and answer space $\mathcal{A}$, we distinguish three objects: the LLM's pre-tool belief $P_x$, the effective tool-conditioned belief $Q_x$, and the authority $\lambda_x$ assigned to the external signal. Here, $P_x,Q_x\in\Delta(\mathcal{A})$, and $\lambda_x\in[0,1]$ controls how far the revised belief moves from $P_x$ toward $Q_x$. The reverse-KL minimal-change objective admits a unique logarithmic-pooling solution:
\begin{equation}
R_x^{\lambda}(a)
\propto
P_x(a)^{1-\lambda_x} Q_x(a)^{\lambda_x},
\qquad a\in\mathcal{A}.
\label{eq:logpool}
\end{equation}
In controlled-output experiments, these beliefs are estimated by projecting answer-probe probabilities onto $\mathcal{A}$.

Collapsing the answer space to the binary event $y^\star$ versus non-gold, let $p=P_x(y^\star)$, $q=Q_x(y^\star)$, and $r=R_x^\lambda(y^\star)$. Denote the corresponding log-odds by $\eta=\mathrm{logit}(p)$, $\nu=\mathrm{logit}(q)$, and $\rho=\mathrm{logit}(r)$. Equation~\eqref{eq:logpool} implies
\begin{equation}
\rho-\eta
=
\lambda_x(\nu-\eta).
\label{eq:logodds}
\end{equation}
Thus, revision is additive in log-odds. Any positive authority decreases the gold log-odds when the tool-conditioned belief is weaker than the internal belief $(\nu<\eta)$, and increases it only when the external support is stronger $(\nu>\eta)$.

\subsection{Protection Principle}

We now characterize when a risk-minimizing policy should grant positive authority to the external signal. For a question with pre-tool gold log-odds $\eta$, let $M$ denote the random gold-event log-odds induced by the tool channel. Define the resulting gold-event logistic risk as
\[
\mathcal{L}(\lambda;\eta)
=
\mathbb{E}_M[\log(1+e^{-z_\lambda(M)})]
\]
where $z_\lambda(M)$ linearly interpolates between the pre-tool and tool-induced gold log-odds.

\begin{theorem}[State-dependent authority threshold]
\label{thm:threshold}
Assume that $M$ is integrable and that all binary gold-event probabilities have finite log-odds. Then $\mathcal{L}(\lambda;\eta)$ is convex in $\lambda$ on $[0,1]$, and its initial slope is
\begin{equation}
\left.\frac{\partial\mathcal{L}}{\partial\lambda}\right|_{\lambda=0}
= -(1-p)\bigl(\mathbb{E}[M]-\eta\bigr).
\label{eq:thresh-deriv}
\end{equation}
Therefore, if $\eta\ge\mathbb{E}[M]$, any positive authority increases risk, and $\lambda^\star=0$ is globally optimal. Conversely, if $\eta<\mathbb{E}[M]$, the initial slope is negative, so a sufficiently small positive authority strictly reduces risk.
\end{theorem}

Theorem~\ref{thm:threshold} establishes a protection principle: when the internal belief already dominates the expected support from the external channel, the minimal-risk policy assigns zero revision authority. Positive external influence is justified only when expected contextual support exceeds the prior.

\paragraph{Empirical Predictions.}
The theory yields three predictions that organize the main empirical tests. \emph{P1:} harm should localize to cases where an internal-belief proxy exceeds a contextual-support proxy. \emph{P2:} harmful revised traces may remain confident, because repeated harmful revision can converge to a sharp but incorrect endpoint (Appendix~\ref{app:proofs:iterated}). \emph{P3:} holding a misleading clue fixed, the revision posture should modulate damage by changing the effective authority assigned to that clue. Section~\ref{sec:mechanism} tests these predictions directly.

\begin{figure*}[!t]
\centering
\includegraphics[width=\textwidth]{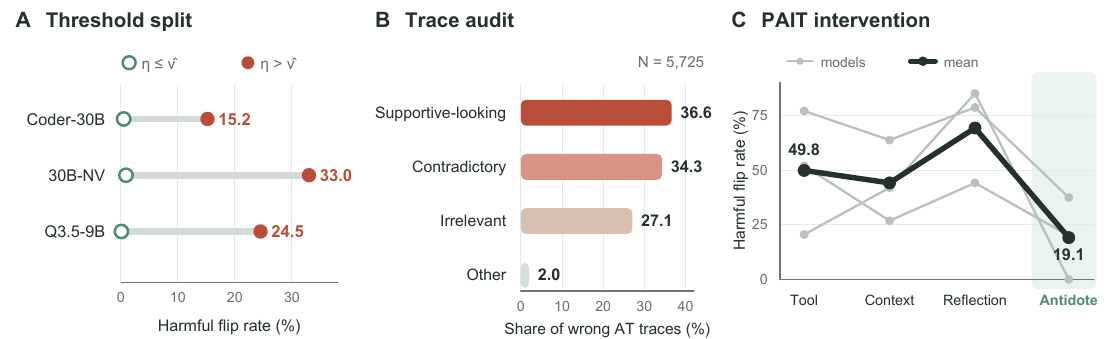}
\caption{\textbf{Diagnostics for misallocated revision authority.}
\textbf{A:} Harmful flips rise sharply in the high-authority region ($\eta>\hat{\nu}$) across three models.
\textbf{B:} Many wrong tool-assisted traces remain superficially plausible, with supportive-looking traces forming the largest category.
\textbf{C:} In PAIT, holding the misleading clue fixed while limiting tool authority reduces harmful flips; gray lines are individual models and the black line is their unweighted mean.}
\label{fig:mechanism_diagnostics}
\end{figure*}

\section{Three Tests of Misallocated Revision Authority}
\label{sec:mechanism}

We next examine whether the threshold mechanism is observable in trace data. Because the latent beliefs in the theory are not directly observed, we test their observable signatures using answer-probe proxies, wrong-trace audits, and controlled revision wrappers, with the concrete measurement mapping reported in Table~\ref{tab:measurement_bridge_appendix}. Throughout this section, NT denotes an unrevised no-tool trajectory, and AT denotes a tool-assisted trajectory after external evidence has been incorporated.

Together, the experiments test three increasingly diagnostic predictions of the theory: harmful flips should localize where internal-belief proxies exceed contextual-support proxies; these failures should often end in confident wrong answers rather than visible uncertainty; holding the same misleading clue fixed, changing the revision frame should change the flip rate by altering the effective authority granted to that clue.

\subsection{Threshold Signature of Harmful Revision}
\label{sec:threshold_signature}

This test asks whether tool-induced harm appears precisely in the regime predicted by the protection principle. The key regime is the \emph{positive-gap} region: cases where the model's pre-tool belief in the gold answer is stronger than the support provided by the tool-conditioned context ($\eta>\hat{\nu}$). In this region, the theory predicts that granting authority to the external signal is risky, because revision is more likely to move probability mass away from the gold answer.

The trace data align with this prediction. In the main Qwen3-Coder-30B export, harmful flips are much more concentrated in the positive-gap region: 15.2\% of these transitions change a correct pre-tool answer into an incorrect post-tool answer, compared with only 0.5\% when the tool-conditioned support is at least as strong as the internal belief ($\eta\le\hat{\nu}$). The average probability update also reverses direction, shifting from a gain of $+0.228$ in $\Delta p_c$ outside the positive-gap region to a loss of $-0.113$ within it. Figure~\ref{fig:mechanism_diagnostics}A visualizes the same split across all three evaluated model settings, while Table~\ref{tab:threshold_transition_main} gives the main-model log-odds and null-crossing diagnostics, with details given in Table~\ref{tab:threshold-cross-model-additional}.

A natural concern is that this effect merely reflects starting confidence: because positive-gap examples begin with high gold probability, their downward updates may produce more boundary crossings simply due to regression to the mean. To address this alternative, we introduce a permutation-style null model \citep{ernst2004permutation,good2005permutation}. Specifically, we keep each example's pre-tool log-odds fixed and randomly reassign the observed logit updates across the dataset. This procedure preserves the marginal distributions of both starting confidence and update magnitude, while severing the case-level link between the threshold gap and the direction of revision.

We observe substantially more crossings than predicted by this null model, especially in the positive-gap and high-prior subsets. In the positive-gap slice, the observed crossing rate is 15.33\% versus 6.09\% under the update-magnitude-matched shuffle; under the stricter dataset-by-difficulty-quartile shuffle it remains 15.33\% versus 10.19\% ($p<.001$ for both). In the high-prior slice, the corresponding rates are 16.32\% versus 5.52\% and 16.32\% versus 10.32\% ($p<.001$). This excess indicates that positive-gap cases are not merely high-confidence or difficult examples with a larger margin for degradation, but rather are disproportionately vulnerable to misaligned revisions. Appendix Table~\ref{tab:matched_null_appendix} gives the complete matched-null audit.

\begin{table}[!t]
\centering
\caption{\textbf{Transition diagnostics for the main Qwen3-Coder-30B threshold risk sets.} 
$\Delta\ell$ is the mean change in gold log-odds. Up/Down report 
logit-movement frequencies, with $|\Delta\ell|<0.05$ treated as flat. 
Cross obs/null reports the observed rate of $p_c>0.5\to p_c<0.5$ 
crossings and the corresponding logit-shuffle null mean.}
\label{tab:threshold_transition_main}
\small
\setlength{\tabcolsep}{3pt}
\resizebox{\linewidth}{!}{%
\begin{tabular}{lrrrrrr}
\toprule
\textbf{Risk set} & \textbf{$N$} & \textbf{$\Delta p_c$} & \textbf{$\Delta\ell$} & \textbf{Up/Down} & \textbf{Harm} & \textbf{Cross obs/null} \\
 & & \textbf{(pp)} & & \textbf{(\%)} & \textbf{(\%)} & \textbf{(\%)} \\
\midrule
$\eta\le\hat{\nu}$ & 70,591 & +22.7 & +4.74 & 69.0 / 25.5 & 0.5 & 0.4 / 0.2 \\
$\eta>\hat{\nu}$ & 13,756 & -11.3 & -1.13 & 44.9 / 53.3 & 15.2 & 15.3 / 6.1 \\
$p_c\ge0.80$ & 11,521 & -15.5 & -1.50 & 43.8 / 54.6 & 15.9 & 16.3 / 5.5 \\
\bottomrule
\end{tabular}%
}
\end{table}

\subsection{Failed Revised Traces Often Appear Superficially Reliable}
\label{sec:silver}

Section~\ref{sec:threshold_signature} localizes where harmful revision occurs. We next ask a complementary question: when a tool-assisted trajectory ends with an incorrect answer, is the failure visible from the trace itself? We audit \emph{AT failures}, i.e., tool-assisted trajectories whose final answer is wrong. This pool includes harmful flips as a strict subset, but the diagnostic target is broader: whether failed revised answer states can be rejected from a single trace. Confidence denotes the final answer-probe probability assigned to the trajectory's own chosen answer; we call a wrong trace high-confidence when this value is at least $\tau_{\mathrm{conf}}=0.8$.

We first annotate each wrong AT trace with an evidence-status label using GPT-5 \citep{openai2025gpt5}: supportive, contradictory, irrelevant, or other/unclassifiable. Figure~\ref{fig:mechanism_diagnostics}B summarizes the audit. First, 36.6\% of wrong AT traces are supportive-looking: the available evidence appears to support the model's final answer, even though that answer is wrong; the remaining 2.0\% are labelled other/unclassifiable. Second, non-supportive evidence does not reliably induce uncertainty: among wrong traces labeled contradictory or irrelevant, 98.2\% still assign at least $\tau_{\mathrm{conf}}=0.8$ probability to their own wrong final answer. Thus, failed revised traces are not reliably exposed by surface confidence or apparent evidence quality alone.

A natural alternative explanation is that these failures are simply bad tool outputs propagated forward. To examine this possibility, we conduct a model-stratified human source audit on 240 wrong AT traces, with 80 traces from each evaluated model. The audit reveals a mixed failure profile: tool-contradictory evidence absorbed anyway is the largest category (34.6\%), followed by irrelevant evidence confabulated into support (24.6\%), partial evidence taken at face value (15.4\%), and correct evidence misinterpreted in the answer space (15.0\%), with 10.4\% labeled unclear. The last category matches the $\binom{85}{82}$ trajectory in the introduction: Python returns the correct value, but the post-tool trajectory revises to the wrong option. Many failures are therefore not artifacts of incorrect tool outputs. Instead, they are post-tool revision failures in which correct, partial, irrelevant, or contradictory evidence can still be folded into a confident wrong state.

Together, the automatic and human audits clarify why threshold localization is not itself a trace-level filter. Although the risky region can be flagged statistically, an individual failed revised trace may still appear supported and highly confident. Apparent evidential support and high final answer-probe confidence are therefore not reliable standalone certificates of a revised trajectory's correctness.

\subsection{Prior-Authority Interaction Test (PAIT)}
\label{sec:pait}

The first two diagnostics establish where harmful revision concentrates and why failed revised traces can look reliable after the fact. We now test the remaining theoretical claim: in high-prior cases, damage should depend not only on the content of the external clue, but also on the authority the model assigns to that clue. In the threshold model, this role is captured by $\lambda$: when the internal belief is already strong, a weak or misleading external signal should receive little or no revision authority. PAIT tests whether this authority is frame-sensitive.

Starting from harmful high-prior cases whose clean executed baseline remains correct and highly confident, we extract one misleading clue and reinsert the same clue into several non-executing wrapper prompts: \textsc{Tool}, \textsc{Context}, \textsc{Reflection}, and \textsc{Tool+Antidote}. The clue text is held fixed within each case, and all wrapper answers disable tool execution. Thus, differences in flip rate cannot be attributed to new tool results or different evidence content; instead, they measure how the same clue is treated under different revision postures. If the harm were driven only by clue content, the frames should behave similarly. If misallocated authority is part of the mechanism, authority-amplifying frames should induce more flips, while an authority-limiting antidote should reduce them.

\begin{table}[t]
\centering
\small
\setlength{\tabcolsep}{4pt}
\caption{\textbf{Cross-model PAIT on eligible high-prior cases.} FR denotes the answer flip rate under the adversarial-clue wrapper.}
\label{tab:pait-cross-model}
\begin{tabular}{lcccc}
\toprule
Model & Tool & Context & Reflection & Tool+Antidote \\
\midrule
QC-30B & 77.0\% & 63.7\% & 78.6\% & 37.4\% \\
30B-NV & 51.9\% & 26.8\% & 44.1\% & 20.0\% \\
Q3.5-9B & 20.5\% & 41.9\% & 85.0\% & 0.0\% \\
\midrule
Avg & 49.8\% & 44.1\% & 69.2\% & 19.1\% \\
\bottomrule
\end{tabular}
\end{table}

Table~\ref{tab:pait-cross-model} summarizes the cross-model PAIT pattern using unweighted model-level averages; the corresponding per-condition eligible counts are reported in Appendix Table~\ref{tab:pait-cross-model-detail}.

The qualitative pattern is stable. Under the same adversarial-clue intervention, multiple authority-coded frames induce substantial harmful revision: averaged across model settings, \textsc{Reflection} produces the highest flip rate (69.2\%), followed by \textsc{Tool} (49.8\%) and \textsc{Context} (44.1\%). At the same time, the revision frame matters: \textsc{Tool+Antidote} yields the lowest average flip rate (19.1\%), suggesting that explicitly limiting the authority of the clue can partially restore protection for high-prior answers.

\section{Deployment Corollary: Authority Allocation for AT/NT Inference}
\label{sec:implication}

The mechanism developed in Sections~\ref{sec:theory}--\ref{sec:mechanism} leads to an inference-time allocation problem. Tool use is valuable because external evidence can rescue weak or mistaken internal states; however, the same revision channel can also overwrite strong priors and produce confident wrong AT endpoints. Once unrevised no-tool trajectories (NT) and revised tool-assisted trajectories (AT) coexist for the same question, the system must decide how much authority each answer state should receive.

This motivates a query-level arbitration view of mixed AT/NT inference. Rather than counting trajectories uniformly, the controller should compare the competing post-revision answer states and allocate additional samples to the channel whose support is weaker or less settled. We instantiate this idea with \textsc{CASE}, a minimal closed-loop controller designed to test the operational prediction of the authority-allocation view: when AT and NT disagree, query-level arbitration should outperform channel-agnostic mixing.

\subsection{A Label-Free Signal from Settled Answer States}
\label{sec:case_signal_main}

The threshold gap $\eta-\mathbb{E}[M]$ relies on leave-one-out tool-conditioned proxies that are not computable per query at deployment time. \textsc{CASE} therefore uses an inference-time analog: the settledness of the resulting answer state from a single trajectory. For each trajectory $i$, we obtain a final answer-probe distribution $p_i(a)$ over options and compute its entropy
\[
H_f^{(i)}=-\sum_{a\in\mathcal{A}}p_i(a)\log p_i(a).
\]
Lower entropy indicates that the trajectory ends in a more concentrated answer state. We convert this quantity into normalized certainty $c_i = 1 - \frac{H_f^{(i)}}{\log |\mathcal{A}|}$
so that $c_i\in[0,1]$ up to numerical roundoff.

\subsection{The \textsc{CASE} Controller}
\label{sec:case_controller_main}

We aggregate answers through certainty-weighted voting. To decide which channel to expand next, \textsc{CASE} also tracks channel-specific winners and margins. For channel $s \in \{\mathrm{AT}, \mathrm{NT}\}$, let $\mathcal{T}_s$ be the current trajectory pool and define
\begin{equation}
\begin{aligned}
V_s(a) &= \sum_{i \in \mathcal{T}_s: y_i = a} c_i,\\
\hat{y}_s &= \arg\max_{a \in \mathcal{A}} V_s(a),\\
\Gamma_s &= V_s(\hat{y}_s) - \max_{b \neq \hat{y}_s} V_s(b).
\end{aligned}
\label{eq:channel_margin}
\end{equation}
Let $\Gamma = V(\hat{y}) - \max_{b \neq \hat{y}} V(b)$ be the corresponding pooled margin. We instantiate these quantities in \textsc{CASE} (\emph{conflict-aware asymmetric expansion}), a minimal closed-loop controller for mixed AT/NT inference.
\begin{algorithm}[!t]
\caption{\textsc{CASE}: conflict-aware asymmetric expansion}
\label{alg:case}
\small
\begin{algorithmic}[1]
\Require Threshold $\tau$; channel budgets $B_{\mathrm{NT}}, B_{\mathrm{AT}}$
\State Collect one NT trajectory and one AT trajectory
\State Initialize $\mathcal{T}_{\mathrm{NT}}, \mathcal{T}_{\mathrm{AT}}$ with those samples
\While{true}
    \State Compute certainty scores $c_i$, pooled winner $\hat{y}$, and pooled margin $\Gamma$
    \State Compute channel winners $\hat{y}_{\mathrm{NT}}, \hat{y}_{\mathrm{AT}}$ and channel margins $\Gamma_{\mathrm{NT}}, \Gamma_{\mathrm{AT}}$
    \If{$\Gamma \ge \tau$ and $\hat{y}_{\mathrm{NT}} = \hat{y}_{\mathrm{AT}}$}
        \State \Return $\hat{y}$
    \ElsIf{$\hat{y}_{\mathrm{NT}} \neq \hat{y}_{\mathrm{AT}}$}
        \State $s^\star \gets \arg\min_{s \in \{\mathrm{NT},\mathrm{AT}\}} V(\hat{y}_s)$
    \Else
        \State $s^\star \gets \arg\min_{s \in \{\mathrm{NT},\mathrm{AT}\}} \Gamma_s$
    \EndIf
    \If{$|\mathcal{T}_{s^\star}| = B_{s^\star}$}
        \State \Return $\hat{y}$
    \EndIf
    \State Collect one additional trajectory from channel $s^\star$ and add it to $\mathcal{T}_{s^\star}$
\EndWhile
\end{algorithmic}
\end{algorithm}

The two branches of Algorithm~\ref{alg:case} correspond to the two findings of Section~\ref{sec:mechanism}. The \emph{conflict branch} handles AT/NT disagreement, the regime in which the threshold gap from Section~\ref{sec:threshold_signature} is most diagnostic: the controller probes whichever side is currently losing the pooled decision rather than reinforcing the stronger one. The \emph{alignment branch} handles cases where both channels agree but the pooled margin remains small---the regime where Section~\ref{sec:silver} shows that surface agreement does not certify correctness---and samples from the channel with the lower margin. Setting $B_{\mathrm{NT}} = B_{\mathrm{AT}} = 1$ yields the minimal one-NT/one-AT form, corresponding to the first decision made by the same closed loop and the setting used for our headline result.

\subsection{Arbitration on the AT/NT Disagreement Set}
\label{sec:case_disagreement_main}

The mechanism in Section~\ref{sec:mechanism} yields a direct operational prediction. When the first AT and NT trajectories choose different answers, equal voting is uninformative by construction; any improvement must therefore come from deciding which post-revision state should be granted authority. We begin with the first-pair AT/NT disagreement subset, which provides the cleanest test of the authority-allocation view.

\begin{table*}[t]
\centering
\caption{\textbf{Cross-model results on the shared four-benchmark evaluation.} Accuracy is measured in the $1{+}1$ setting with one NT and one AT trajectory per question. Baselines are mixed majority vote with random tie-breaking and P(True)-CISC on the same candidate files. The final row reports matched-budget Pareto AUC over $1{+}1$ to $8{+}8$; $\Delta_{\mathrm{C}}$ denotes \textsc{CASE}--P(True)-CISC.}
\label{tab:minimal_result}
\small
\setlength{\tabcolsep}{3.0pt}
\resizebox{\textwidth}{!}{%
\begin{tabular}{lcccccccccccc}
\toprule
\textbf{Dataset / summary} & \multicolumn{4}{c}{\textbf{Qwen3-Coder-30B}} & \multicolumn{4}{c}{\textbf{30B-NV}} & \multicolumn{4}{c}{\textbf{Qwen3.5-9B}} \\
\cmidrule(lr){2-5}\cmidrule(lr){6-9}\cmidrule(lr){10-13}
& \textbf{Mix-MV} & \textbf{CISC} & \textbf{\textsc{CASE}} & \textbf{$\Delta_{\mathrm{C}}$} & \textbf{Mix-MV} & \textbf{CISC} & \textbf{\textsc{CASE}} & \textbf{$\Delta_{\mathrm{C}}$} & \textbf{Mix-MV} & \textbf{CISC} & \textbf{\textsc{CASE}} & \textbf{$\Delta_{\mathrm{C}}$} \\
\midrule
MMLU-Pro & 73.95 & 73.91 & 78.29 & +4.38 & 82.76 & 82.93 & 84.31 & +1.38 & 60.71 & 56.65 & 62.90 & +6.25 \\
GPQA Diamond & 48.21 & 51.79 & 49.74 & -2.05 & 66.16 & 65.66 & 67.17 & +1.51 & 62.88 & 65.15 & 63.64 & -1.51 \\
Physics & 48.46 & 50.31 & 50.62 & +0.31 & 69.08 & 69.54 & 69.85 & +0.31 & 58.31 & 56.92 & 58.15 & +1.23 \\
SuperGPQA-Calc & 42.92 & 42.38 & 46.44 & +4.06 & 56.91 & 57.82 & 59.96 & +2.14 & 34.38 & 31.43 & 36.35 & +4.92 \\
\midrule
Macro & 53.38 & 54.60 & 56.27 & +1.67 & 68.73 & 68.99 & 70.32 & +1.33 & 54.07 & 52.54 & 55.26 & +2.72 \\
Aggregate & 53.55 & 53.27 & 57.30 & +4.03 & 65.97 & 66.62 & 68.46 & +1.84 & 44.04 & 40.80 & 46.03 & +5.23 \\
\textit{Pareto AUC} & 57.48 & 57.31 & 59.69 & +2.38 & 68.94 & 69.12 & 70.63 & +1.51 & 46.49 & 44.96 & 48.17 & +3.21 \\
\bottomrule
\end{tabular}
}%
\end{table*}
A pair-complete $1$-vs.-$1$ audit provides the uncertainty qualification requested during review. Over 53,184 matched model--question instances, a model- and dataset-stratified paired bootstrap gives \textsc{CASE}--P(True)-CISC $=+3.80$ points with a 95\% interval of $[3.54,4.05]$. The model-level intervals all exclude zero, whereas the pooled GPQA Diamond contrast is $0.00$ with interval $[-2.37,2.37]$; GPQA is therefore inconclusive rather than evidence of a reliable reversal. Appendix Table~\ref{tab:case_bootstrap_appendix} reports the full breakdown.

Table~\ref{tab:case_disagreement_main} reports the complementary disagreement-only test. On this subset, \textsc{CASE} improves over P(True)-CISC by $+10.2$, $+10.0$, and $+13.1$ points across the three full-pipeline model settings.

The decomposition explains the source of the gain. \textsc{CASE} does not simply default to either the no-tool channel or the tool-assisted channel: it protects 60--81\% of cases where NT is correct and AT is wrong, while still rescuing 51--71\% of cases where only AT is correct. This pattern is consistent with the threshold law, indicating that the gains arise from state-dependent authority allocation rather than from a uniform preference for either the tool-assisted or no-tool channel.

\begin{figure*}[!t]
\centering
\includegraphics[width=\textwidth]{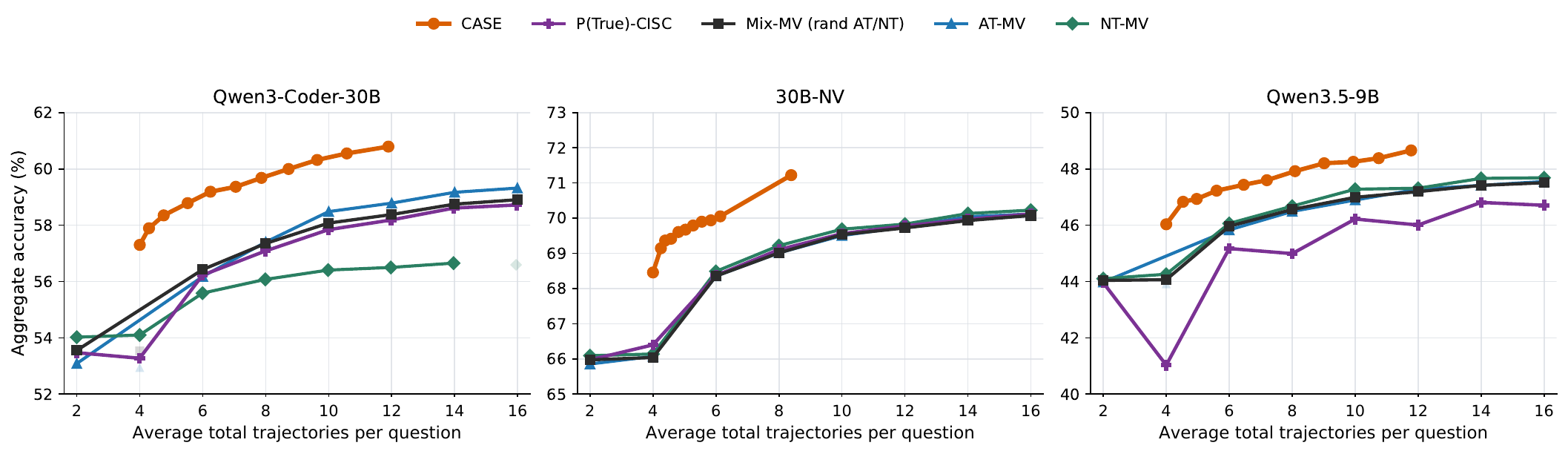}
\caption{\textbf{\textsc{CASE} versus majority-vote baselines on the shared four-benchmark evaluation.} Each panel corresponds to one model setting. The x-axis reports total sampled trajectories per question rather than token-normalized cost. \textsc{Mix-MV} samples AT/NT trajectories uniformly and applies equal-weight majority voting with uniform tie expectation. Solid lines are Pareto frontiers; faint markers show all evaluated operating points. \textsc{CASE} instead uses the authority signal to allocate AT versus NT samples.}
\label{fig:case_pareto}
\end{figure*}

\begin{table}[t]
\centering
\caption{\textbf{First-pair AT/NT disagreement decomposition.} Accuracy is measured only on examples where the first NT and AT answers disagree. False protection is the complement of rescue on the NT-wrong/AT-correct subset.}
\label{tab:case_disagreement_main}
\scriptsize
\setlength{\textwidth}{2.2pt}
\begin{tabular}{lrrrrrr}
\toprule
\textbf{Model} & \textbf{$N$} & \textbf{CASE} & \textbf{CISC} & \textbf{Prot.} & \textbf{Rescue} & \textbf{False} \\
\midrule
Qwen3-Coder & 6{,}975 & 37.7 & 27.5 & 81.3 & 51.1 & 48.9 \\
30B-NV & 3{,}897 & 39.2 & 29.2 & 71.2 & 70.5 & 29.5 \\
Qwen3.5-9B & 6{,}905 & 34.0 & 20.9 & 60.4 & 57.4 & 42.6 \\
\bottomrule
\end{tabular}
\end{table}

Figure~\ref{fig:case_pareto} examines whether the same arbitration advantage persists beyond the minimal-pair setting. Across the $1{+}1$ to $8{+}8$ budget ladder, \textsc{CASE} remains on a better budget frontier than mixed majority voting and CISC. Detailed cost tables, prior-gate baselines, and nearby answer-state signals are reported in Appendix~\ref{app:case}.

To further test whether \textsc{CASE}'s advantage concentrates in the regimes predicted by the threshold analysis, we stratify the disagreement set by $\eta>\hat\nu$ versus $\eta\le\hat\nu$ and by prior strength. Averaged across model settings, the \textsc{CASE}--CISC gap is larger in the positive-gap region than outside it ($+13.5$ vs. $+7.7$), and it reaches $+20.8$ points in the high-prior bin; Appendix Table~\ref{tab:case_strata_support} reports the corresponding stratified summaries. In that high-prior bin, the gain is not explained by a degenerate rule that always trusts NT: \textsc{CASE} protects 86.5\% of NT-correct/AT-wrong cases while still rescuing 72.5\% of NT-wrong/AT-correct cases. This stratification closes the loop between the mechanism and the controller. \textsc{CASE} does not observe the latent threshold gap, but its advantage concentrates precisely in the regimes where the threshold analysis predicts over-revision. 

\subsection{Summary}
Section~\ref{sec:implication} shows that the belief-revision mechanism has a direct operational consequence. Since the latent threshold gap is unavailable at inference time, \textsc{CASE} uses answer-state settledness as a label-free authority cue between competing AT and NT trajectories. Its gains concentrate in the disagreement and high-prior regimes predicted by the earlier analysis, while retaining both protection and rescue across minimal setting and higher budgets. This supports the main reframing of mixed tool/no-tool inference as state-dependent authority allocation rather than uniform trajectory counting.

\section{Related Work}
\label{sec:related}

\paragraph{External evidence, tool use, and belief revision.}
Tool-integrated reasoning extends language models beyond parametric
computation through code execution, retrieval, and search, increasingly
trained end-to-end with reinforcement learning and standardized
evaluation
\citep{yao2023react,schick2023toolformer,paranjape2023art,wang2024codeact,
zhuang2023toolqa,feng2025retool,jin2025searchr1,lin2025understanding}.
A growing body of work shows that external evidence is not uniformly
beneficial: irrelevant or distracting context degrades accuracy, long
contexts are poorly utilized, retrieval-augmented systems over-adopt
incorrect evidence, and agents may overuse tools even when parametric
reasoning suffices
\citep{shi2023large,liu2024lost,yoran2024irrelevant,mallen2023when,
wu2024clasheval,qian2025smart}.
A parallel literature on knowledge conflict characterizes the tug-of-war
between parametric memory and external context---disentangling
parametric from contextual answers, surfacing entity- and
context-faithfulness conflicts, quantifying internal--external
trade-offs, and cataloguing conflict types at scale
\citep{neeman2023disentqa,longpre2021entity,zhou2023contextfaithful,
wu2024clasheval,xie2024adaptive,su2024conflictbank,
kortukov2024conflicts,xu2024knowledge}.
Mitigation strategies span source allocation, conflict-specific decoding,
and context-robust training objectives
\citep{tao2024context,jin2024tug,jin2024cutting,wang2025astuterag,
zeng2025contextrobust,khandelwal2025cocoa}.
We build on this literature but shift the unit of analysis from retrieved
passages or tool-use decisions to tool-returned reasoning traces, asking
when a tool-conditioned state should receive zero revision authority
because the pre-tool belief is already stronger.

\paragraph{Aggregation and adaptive inference.}
Our deployment result connects to test-time aggregation and adaptive
computation. Chain-of-thought prompting and self-consistency aggregate
sampled reasoning paths, while confidence-weighted variants leverage
calibration, self-knowledge, semantic uncertainty, and self-checking
signals to score or filter candidates
\citep{wei2022chain,kojima2022large,wang2022self,taubenfeld2025cisc,
guo2017calibration,jiang2021calibration,kadavath2022language,
farquhar2024semantic,manakul2023selfcheckgpt,tan2025consistent}.
Verifier and process-supervision methods offer a complementary route to
selecting or training reasoning paths via outcome- and step-level reward
models
\citep{cobbe2021training,uesato2022solving,lightman2024lets,
zelikman2022star}.
Adaptive systems further decide when to retrieve, decode, or route
computation across models and contexts
\citep{asai2024selfrag,khandelwal2025cocoa,chen2023frugalgpt}.
CASE is complementary to these directions: rather than training a
verifier or routing among models, it tests the operational implication
of our mechanism by allocating authority between unrevised no-tool and
revised tool-assisted trajectories within a fixed mixed budget.

\section{Conclusion}
Our results show that tool access changes not only the information available to a model, but also the authority
external evidence holds over an existing belief. Across trace diagnostics,
controlled interventions, and arbitration experiments, tool-induced harm is
consistently state dependent. The same revision channel that rescues weak
beliefs can overwrite strong correct ones, and the damage concentrates when
internal support already exceeds the expected support of the tool channel.
Therefore, tool use be understood as a belief-revision intervention
whose value depends on the model's current answer state, rather than
unconditional capability. \textsc{CASE} operationalizes this view at inference
time by protecting states in which external evidence is predicted to be more
harmful than helpful. More broadly, it suggests a mechanism-level objective for
tool learning, where models learn to allocate revision authority according
to the relative support of internal and tool-conditioned evidence, rather than
tool reliability alone.
\section*{Limitations}
\label{sec:limitations}

Our study has several limitations. The empirical evidence is concentrated on a bounded set of model families, benchmarks, and tool-use formats. Although we evaluate multiple model settings and four reasoning benchmarks, broader validation on larger frontier-scale models and more diverse architectures remains necessary. The current experiments focus primarily on controlled tool-assisted reasoning traces, where the external channel and answer space can be probed and compared. More complex tool-use settings, such as coding agents, web-search agents, retrieval-augmented multi-hop systems, and long-horizon autonomous agents, may introduce additional sources of noise, delayed feedback, and compounding state changes. Whether the same threshold behavior appears in these richer environments, and how to design a corresponding \textsc{CASE}-style controller for them, remains an important direction for future work.

For deployment of the core finding, \textsc{CASE} should be interpreted as a minimal operational corollary rather than a complete deployment solution. We do not yet show how the protection principle should be incorporated into training. A natural next step is to use the state-dependent authority view to design supervised fine-tuning data, preference signals, or reinforcement-learning rewards that teach models when to preserve a strong internal belief, when to defer to external evidence, and when to request additional tool interaction. Integrating such objectives with modern supervised fine-tuning and reinforcement-learning pipelines for tool-integrated reasoning and LLM agents remains open.

\section*{Ethics Statement}
\label{sec:ethics}

The experiments use benchmark reasoning tasks and stored model/tool traces rather than human-subject data. The main risk raised by the paper is operational: tool-augmented systems can become confidently wrong when misleading or low-quality external evidence is granted too much revision authority. We therefore frame the proposed controller as a safety-motivated diagnostic and mitigation direction. Deployed systems should combine state-dependent authority allocation with independent tool-quality checks and provenance tracking; high-stakes domains also require task-specific human oversight.  Experts involved in the annotation process were compensated in accordance with applicable regulations and institutional guidelines.

\section*{Acknowledgment}
This research was supported by the National Natural Science Foundation of China under Grant 62271285.

\bibliography{references}

\clearpage
\appendix
\raggedbottom
\setcounter{table}{0}
\setcounter{figure}{0}
\makeatletter
\@addtoreset{table}{section}
\@addtoreset{figure}{section}
\makeatother
\renewcommand{\thetable}{\thesection.\arabic{table}}
\renewcommand{\thefigure}{\thesection.\arabic{figure}}

\section{Mechanism Measurements and Robustness}
\label{app:measurement}

\subsection{Theory-to-measurement bridge}
\label{sec:measurement_bridge}

For multiple-choice runs, pre-tool and post-tool states are reconstructed from the fixed answer-space probe attached to each stored trajectory. Thus $p_{c,\mathrm{pre}}$, $p_{c,\mathrm{post}}$, and $\Delta p_c=p_{c,\mathrm{post}}-p_{c,\mathrm{pre}}$ come from the same probe family rather than from free-form textual heuristics.

\begin{table}[H]
\centering
\caption{\textbf{Consolidated notation and empirical proxies.} The first rows are theory objects; hatted quantities are observed diagnostics used for localization and sign tests, not identified latent parameters.}
\label{tab:measurement_bridge_appendix}
\footnotesize
\setlength{\tabcolsep}{3pt}
\renewcommand{\arraystretch}{1.08}
\begin{tabularx}{\linewidth}{L{0.29\linewidth}YY}
\toprule
\textbf{Quantity} & \textbf{Definition or proxy} & \textbf{Role} \\
\midrule
$P_x,Q_x,R_x^\lambda$ & Pre-tool, tool-conditioned, and revised beliefs over $\mathcal{A}$ & Full answer-space states \\
$p,q,r$ & $P_x(y^\star)$, $Q_x(y^\star)$, and $R_x^\lambda(y^\star)$ after collapsing to gold vs. non-gold & Binary gold-event probabilities \\
$\eta,\nu,\rho$ & $\operatorname{logit}(p)$, $\operatorname{logit}(q)$, and $\operatorname{logit}(r)$ & Binary log-odds coordinates \\
$M$, $\bar\nu$ & Random tool-induced gold log-odds $M$ and expected support $\bar\nu=\mathbb{E}[M]$ & Threshold in Theorem~\ref{thm:threshold} \\
$\hat q_x$, $\hat\nu_x$ & Leave-one-out matched post-tool gold probability $\hat q_x$ and $\hat\nu_x=\operatorname{logit}(\operatorname{clip}_\epsilon(\hat q_x))$ & Empirical proxy for $\bar\nu$ \\
$\lambda_x$, $\hat\lambda_x$ & Latent revision authority and diagnostic recovered-authority ratio in Eq.~\eqref{eq:lambda_hat_diag} & Measures observed movement toward the proxy support \\
$H_f$, $c_i$ & Final answer entropy and normalized certainty $c_i=1-H_f/\log|\mathcal{A}|$ & Label-free \textsc{CASE} arbitration signal \\
\bottomrule
\end{tabularx}
\end{table}

For numerical stability, all logits use $\operatorname{clip}_\epsilon(u)=\min(1-\epsilon,\max(\epsilon,u))$ with $\epsilon=10^{-6}$. For an event $x$, let $\mathcal{S}_x$ be the matched stratum \texttt{(dataset, tool\_call\_idx, prior\_bucket, pre\_correct)}. The main contextual-support proxy is
\begin{equation}
\begin{aligned}
\hat q_x
&=\frac{1}{|\mathcal{S}_x|-1}
  \sum_{j\in\mathcal{S}_x,\ j\ne x} p_{c,\mathrm{post}}^{(j)},\\
\hat\nu_x
&=\operatorname{logit}(\operatorname{clip}_\epsilon(\hat q_x)).
\end{aligned}
\label{eq:nu_hat_proxy}
\end{equation}
This definition is used for retained events with a nonempty leave-one-out stratum; robustness widens the strata and uses cross-fit or oracle-support variants. Let $\rho_x=\operatorname{logit}(\operatorname{clip}_\epsilon(p_{c,\mathrm{post}}^{(x)}))$ and $\eta_x=\operatorname{logit}(\operatorname{clip}_\epsilon(p_{c,\mathrm{pre}}^{(x)}))$. The recovered-authority diagnostic reported in Table~\ref{tab:threshold-cross-model-additional} is
\begin{equation}
\hat\lambda_x=
\operatorname{clip}_{[0,1]}
\left(\frac{\rho_x-\eta_x}{\hat\nu_x-\eta_x}\right),
\quad |\hat\nu_x-\eta_x|>10^{-6}.
\label{eq:lambda_hat_diag}
\end{equation}
with near-zero denominators treated as undefined and excluded from the diagnostic mean. Because effective authority is latent and $\hat\nu_x$ is only a proxy for expected channel support, $\hat\lambda_x$ is used only as a descriptive geometry check. The key empirical claim is sign stability of the belief-minus-support gap across proxy families.

\subsection{Trace-level checks}
\label{sec:qualitative_cases}

\paragraph{Executed AT failure.}
On the SuperGPQA-Calc binomial item described in the introduction, the pre-tool probe assigns $p_c=0.99995$ to the combined correct option. Python returns \texttt{C(85,82)=98770}, \texttt{C(85,3)=98770}, and equality \texttt{True}; the tool-assisted trajectory still answers the numeric-only option. This case illustrates that the external computation need not be wrong, since harm can come from over-authorizing a revised interpretation of the answer space. To construct the subsequent PAIT experiment, human annotators rewrite the misleading information in each stored harmful AT trajectory into a short, wrapper-invariant clue and insert that same clue verbatim across the PAIT wrappers described in Appendix~\ref{sec:pait_protocol}.

\paragraph{\textsc{CASE} arbitration.}
On an MMLU-Pro fraction item, the first AT path calls Python and observes \texttt{Result: 0.625} but answers a wrong option with a split final probe; the first NT path answers the gold option with near-zero entropy. Equal voting ties, while \textsc{CASE}(1+1) gives the larger authority weight to the more settled NT answer state.

\subsection{Asymmetric belief destruction by dataset}
\label{sec:abd}

This subsection records the compact numerical evidence behind the per-dataset visual diagnostics. The corresponding full-width figures are postponed to Section~\ref{sec:appendix_figures} so that the appendix first presents the definitions, trace checks, and tables without a float backlog.

Let $\Delta p_c^+=\max(\Delta p_c,0)$ and $\Delta p_c^-=\min(\Delta p_c,0)$. Across datasets, the expected magnitude of harmful updates grows faster than the expected magnitude of helpful ones as prior confidence increases:
\begin{equation}
\frac{\mathbb{E}[|\Delta p_c^-|\mid p_c]}
{\mathbb{E}[\Delta p_c^+\mid p_c]}
=
A\left(\frac{p_c}{1-p_c}\right)^\alpha .
\label{eq:destruction_law}
\end{equation}
This descriptive relation is the precursor to the threshold analysis: it summarizes how the harm/help balance shifts with the model's current belief state. A positive exponent means that moving to higher pre-tool odds increases the relative size of destructive updates. Values near one indicate an approximately odds-proportional increase in destructive amplitude.

\begin{table}[H]
\centering
\caption{\textbf{Destructive-asymmetry fits.} The exponent $\alpha$ is positive across the four main benchmarks.}
\label{tab:acs_params}
\footnotesize
\setlength{\tabcolsep}{4pt}
\renewcommand{\arraystretch}{1.04}
\begin{tabular}{lrr}
\toprule
\textbf{Dataset} & \textbf{$\alpha$} & \textbf{$R^2$} \\
\midrule
MMLU-Pro & 1.075 & 0.973 \\
GPQA Diamond & 0.960 & 0.989 \\
Physics & 0.997 & 0.995 \\
SuperGPQA-Calc & 1.092 & 0.975 \\
\bottomrule
\end{tabular}
\end{table}

Table~\ref{tab:acs_params} shows that the fitted exponents are positive in all four datasets and tightly fit the binned ratios. As the pre-tool gold probability increases, harmful updates become larger relative to helpful updates across MMLU-Pro, GPQA Diamond, Physics, and SuperGPQA-Calc.

\begin{table}[H]
\centering
\caption{\textbf{Cross-model portability summary.} Destructive-asymmetry exponents remain positive, and \textsc{CASE}(1+1) improves aggregate accuracy in all three evaluated settings.}
\label{tab:cross_model_scope}
\footnotesize
\setlength{\tabcolsep}{3pt}
\renewcommand{\arraystretch}{1.04}
\begin{tabular}{lrrrr}
\toprule
\textbf{Model} & \textbf{Events} & \textbf{$\bar\alpha$} & \textbf{$\bar R^2$} & \textbf{CASE $\Delta$} \\
\midrule
Qwen3-Coder-30B & 84,347 & 1.031 & 0.983 & +3.8 \\
30B-NV & 149,147 & 0.646 & 0.925 & +2.5 \\
Qwen3.5-9B & 71,028 & 1.047 & 0.986 & +2.0 \\
\bottomrule
\end{tabular}
\end{table}

Table~\ref{tab:cross_model_scope} adds two portability checks. First, the destructive-asymmetry exponent remains positive for each evaluated model setting, although the slope is shallower for 30B-NV. Second, the same settings are those in which \textsc{CASE} improves one-NT/one-AT aggregate accuracy, linking the descriptive asymmetry to the deployment-time arbitration result.

\subsection{Threshold-signature robustness}
\label{sec:proxy_robustness}

The main threshold split uses an effective support proxy. We therefore report the same split in log-odds coordinates and stress-test the sign of the gap under alternative proxy families. For numerical stability, log-odds are computed after clipping $p_c$ to $[10^{-6},1-10^{-6}]$.

\begin{table}[H]
\centering
\caption{\textbf{Boundary-calibrated transition diagnostics.} Crossings are $p_c>0.5\to p_c<0.5$; the null shuffles empirical logit updates across events.}
\label{tab:boundary_transition_full}
\scriptsize
\setlength{\tabcolsep}{2.5pt}
\renewcommand{\arraystretch}{1.03}
\resizebox{\linewidth}{!}{%
\begin{tabular}{lrrrrrr}
\toprule
\textbf{Risk set} & \textbf{$N$} & \textbf{$\Delta p_c$} & \textbf{$\Delta\ell$} & \textbf{Down} & \textbf{Harm} & \textbf{Cross/null} \\
 & & \textbf{(pp)} & & \textbf{(\%)} & \textbf{(\%)} & \textbf{(\%)} \\
\midrule
All events & 84,347 & +17.2 & +3.78 & 30.1 & 2.9 & 2.8 / 1.2 \\
Trimmed $0.05\le p_c\le0.95$ & 7,000 & +7.1 & +0.97 & 44.7 & 14.3 & 12.9 / 8.6 \\
$p_c\ge0.80$ & 11,521 & -15.5 & -1.50 & 54.6 & 15.9 & 16.3 / 5.5 \\
$p_c\ge0.95$ & 10,146 & -14.8 & -1.74 & 55.6 & 14.2 & 14.4 / 4.0 \\
$\eta>\hat\nu$ & 13,756 & -11.3 & -1.13 & 53.3 & 15.2 & 15.3 / 6.1 \\
$\eta\le\hat\nu$ & 70,591 & +22.7 & +4.74 & 25.5 & 0.5 & 0.4 / 0.2 \\
\bottomrule
\end{tabular}%
}
\end{table}

Table~\ref{tab:boundary_transition_full} isolates the same pattern in several risk sets. The positive-gap slice $\eta>\hat\nu$ has negative mean probability and logit updates, a much higher downward-movement frequency, and an observed crossing rate that is well above the logit-shuffle null. The high-prior slices show the same qualitative pattern, which is the empirical signature predicted by the threshold law.

\begin{table}[H]
\centering
\caption{\textbf{Threshold signature across proxy choices.} $\Delta$harm is the high-gap minus low-gap harmful-flip rate in percentage points.}
\label{tab:proxy_robustness}
\scriptsize
\setlength{\tabcolsep}{2.5pt}
\renewcommand{\arraystretch}{1.03}
\begin{tabularx}{\linewidth}{L{0.22\linewidth}L{0.28\linewidth}r r L{0.16\linewidth}}
\toprule
\textbf{Belief} & \textbf{Support} & \textbf{$N$} & \textbf{$\Delta$harm} & \textbf{95\% CI} \\
\midrule
Pre-tool $p_c$ & Overlap, cross-fit & 100,919 & +11.1 & [10.6, 11.5] \\
Pre-tool $p_c$ & Overlap & 100,919 & +10.3 & [9.9, 10.7] \\
NT mean $p_c$ & Overlap, cross-fit & 47,762 & +10.2 & [9.6, 10.9] \\
NT mean $p_c$ & Overlap & 47,762 & +9.5 & [8.9, 10.1] \\
NT mean $p_c$ & Oracle support & 44,128 & +6.4 & [6.0, 6.9] \\
Pre-tool $p_c$ & Oracle support & 93,016 & +6.1 & [5.8, 6.4] \\
\bottomrule
\end{tabularx}
\end{table}

Across all six proxy pairs in Table~\ref{tab:proxy_robustness}, the positive-gap regime is more harmful. The harm penalty ranges from +6.1 to +11.1 points, and every bootstrap interval excludes zero. Under the main pre-tool/overlap proxy, the raw unbinned relation is also negative: OLS slope $-0.013$, $R^2=0.066$, and Spearman $\rho=-0.194$ over 100,919 events.

\subsection{Cross-model threshold diagnostics}
\label{app:threshold-cross-model}

\begin{table*}[!t]
\centering
\caption{\textbf{Detailed threshold diagnostics for the additional model settings.} Rows $\eta\le\hat\nu$ and $\eta>\hat\nu$ form the threshold split; $p_{c,\mathrm{pre}}\ge0.80$ is a high-prior robustness slice rather than a disjoint partition. Harm and Rescue are flip rates. $\hat\lambda$ is a diagnostic recovered-authority estimate.}
\label{tab:threshold-cross-model-additional}
\footnotesize
\setlength{\tabcolsep}{3.0pt}
\resizebox{\textwidth}{!}{%
\begin{tabular}{@{}llrrrrrrrr@{}}
\toprule
\textbf{Model} & \textbf{Slice} & \textbf{$N$} & \textbf{Harm} & \textbf{Rescue} & \textbf{Mean $\Delta p_c$} & \textbf{Mean $\Delta\ell$} & \textbf{Up/Flat/Down} & \textbf{Mean $\eta-\hat\nu$} & \textbf{Mean $\hat\lambda$} \\
\midrule
30B-NV & $\eta\le\hat\nu$ & 96,636 & 0.9 & 1.3 & +0.0102 & +0.722 & 20.3/71.9/7.8 & -0.0257 & 0.106 \\
30B-NV & $\eta>\hat\nu$ & 52,511 & 33.0 & 4.3 & -0.2909 & -9.075 & 26.9/3.7/69.4 & +0.3195 & 0.605 \\
30B-NV & $p_{c,\mathrm{pre}}\ge0.80$ & 20,359 & 73.4 & 0.0 & -0.7042 & -16.021 & 7.5/2.4/90.2 & +0.7042 & 0.782 \\
\addlinespace
Qwen3.5-9B & $\eta\le\hat\nu$ & 34,724 & 0.1 & 5.4 & +0.0398 & -5.061 & 25.7/29.2/45.1 & -0.0589 & 0.134 \\
Qwen3.5-9B & $\eta>\hat\nu$ & 36,304 & 24.5 & 5.9 & -0.1956 & -4.304 & 22.7/1.1/76.2 & +0.2139 & 0.589 \\
Qwen3.5-9B & $p_{c,\mathrm{pre}}\ge0.80$ & 21,663 & 30.5 & 0.0 & -0.2898 & -3.000 & 19.4/0.6/80.0 & +0.2898 & 0.499 \\
\bottomrule
\end{tabular}%
}
\end{table*}

Table~\ref{tab:threshold-cross-model-additional} expands the compact cross-model summary from Section~\ref{sec:threshold_signature} for the two additional model settings. Together with the main-export diagnostics in Table~\ref{tab:boundary_transition_full}, it shows that the $\eta>\hat\nu$ region has higher harmful-flip rates and more negative probability updates across evaluated settings.

The table also explains why the paper emphasizes the paired harmful-flip and probability-update signature rather than any single logit-space statistic. Probability-space and logit-space summaries can differ in low-gap slices because the logit transform amplifies boundary movements, especially near the simplex boundary. Even with this caveat, the high-gap rows consistently show higher recovered authority and larger destructive movement, matching the state-dependent revision mechanism.

\subsection{Per-dataset analysis on tool revision mechanisms}
\label{sec:appendix_figures}

\begin{figure*}[p]
  \centering
  \includegraphics[width=\textwidth]{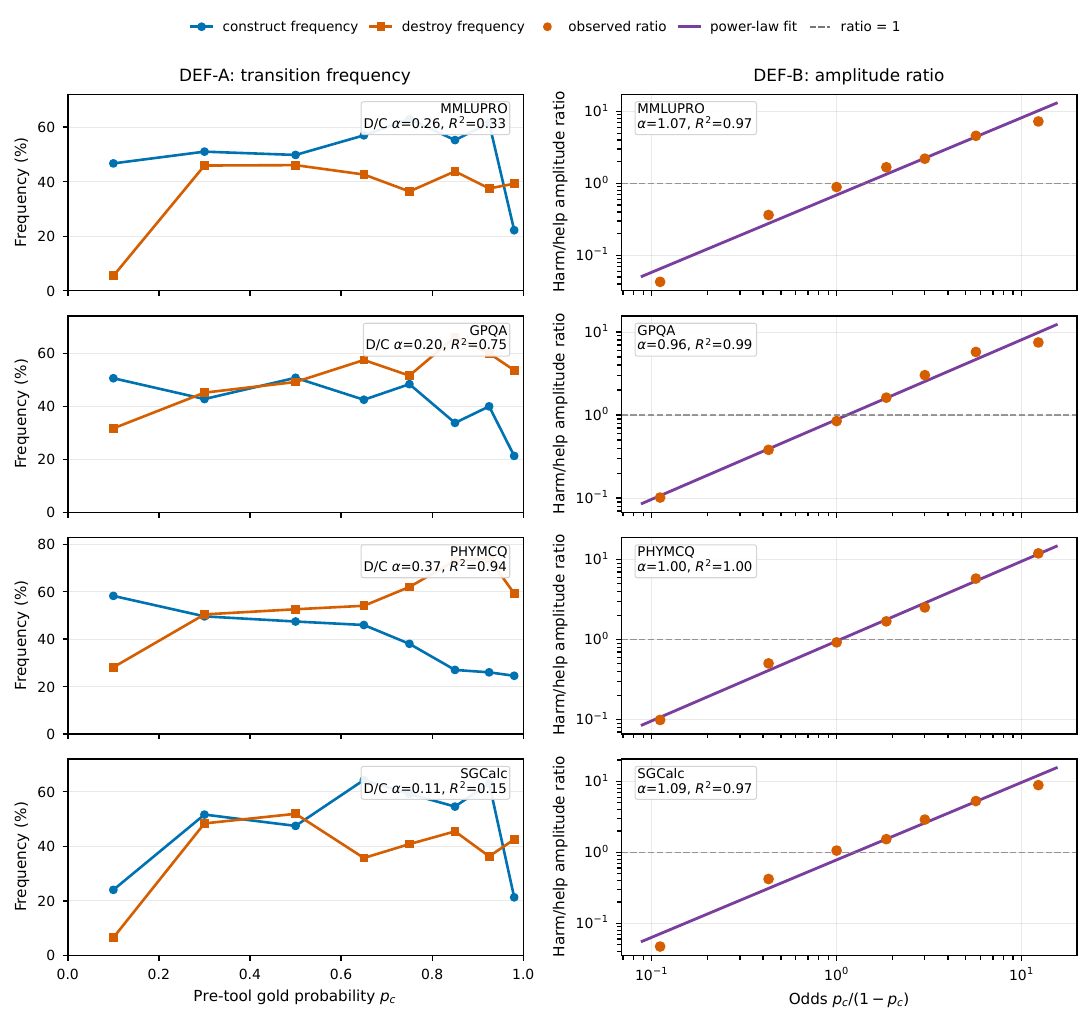}
  \caption{\textbf{Per-dataset diagnostics for asymmetric belief destruction.} Each row is one benchmark. The left column separates constructive transitions, which increase gold probability, from destructive transitions, which decrease it, across pre-tool gold-probability bins. The right column plots the harmful/helpful amplitude ratio $\mathbb{E}[|\Delta p_c^-|\mid p_c] / \mathbb{E}[\Delta p_c^+\mid p_c]$ against pre-tool odds on log--log axes. The fitted positive slopes show that as the model enters stronger-prior states, destructive update amplitude grows faster than constructive update amplitude. The horizontal ratio-$1$ reference line marks the point where helpful and harmful update magnitudes are equal; points above it are harm-dominant.}
  \label{fig:appendix-frequency-amplitude}
\end{figure*}

\begin{figure*}[p]
  \centering
  \includegraphics[width=\textwidth]{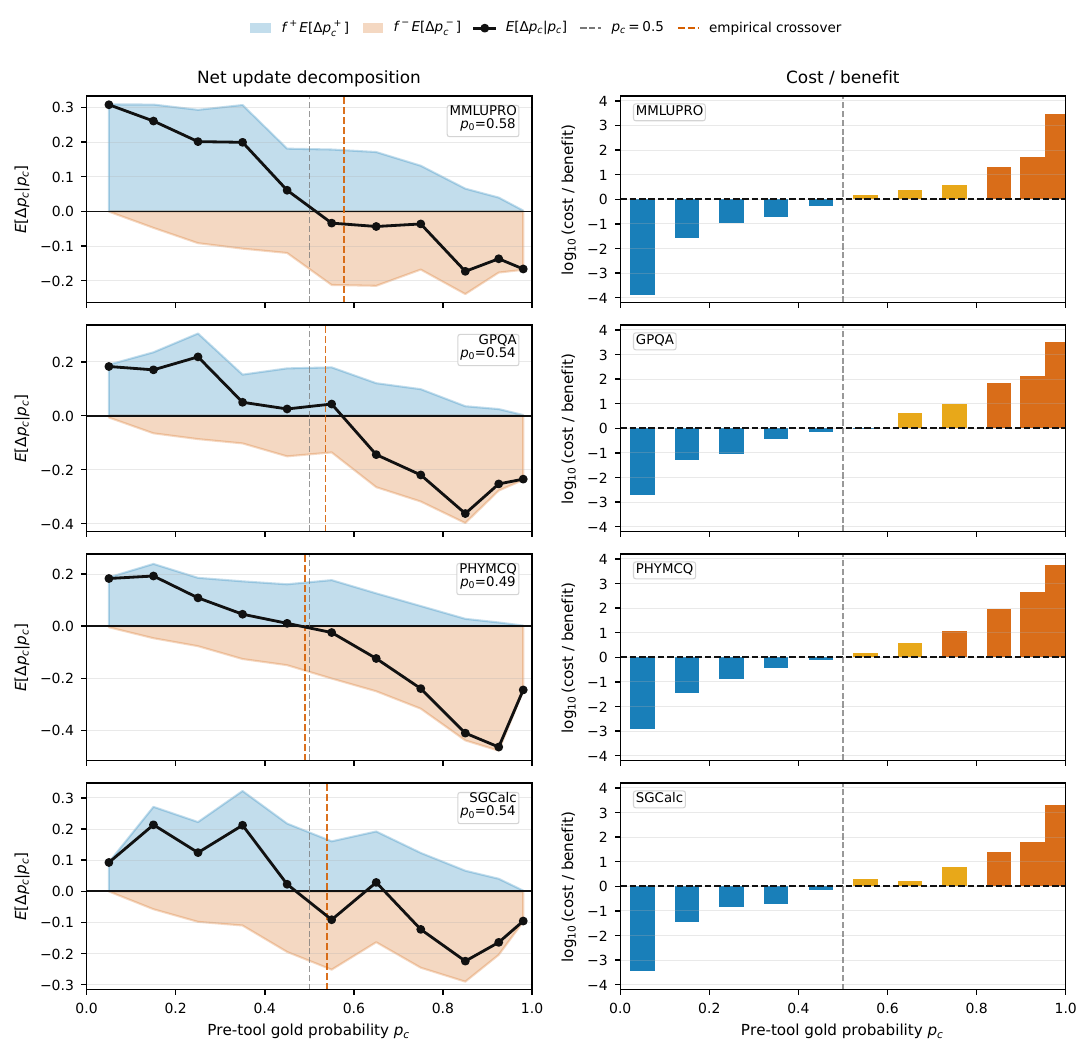}
  \caption{\textbf{Net effect of asymmetric belief revision.} The left column decomposes $\mathbb{E}[\Delta p_c\mid p_c]$ into helpful and harmful contributions. The black net-update curve shows where tool use changes from helpful on average to harmful on average; vertical dashed lines mark $p_c=0.5$ and the empirical zero crossing. The right column expresses the same shift as a cost/benefit ratio on a $\log_{10}$ scale. Bars above zero indicate bins where harmful update mass exceeds helpful update mass, so the external channel has become net destructive despite being useful at lower priors.}
  \label{fig:appendix-delta-cost}
\end{figure*}

Figure~\ref{fig:appendix-frequency-amplitude} answers whether the destructive pattern is driven by event frequency or by update size. Read it row-wise: each row fixes a benchmark, so the left and right panels in that row describe the same set of binned transitions. The left panel is a frequency view. It asks how often tool use moves probability in the helpful direction versus the harmful direction at each pre-tool probability bin. These frequencies alone do not prove net harm, because a rare destructive move can be large and a frequent constructive move can be small. They are nevertheless useful for ruling out a trivial story in which high-prior harm appears only because destructive transitions are vanishingly rare but heavily weighted by a few outliers.

The right panel of Figure~\ref{fig:appendix-frequency-amplitude} is the amplitude view, and it is the cleaner diagnostic for asymmetric destruction. The x-axis converts $p_c$ to prior odds $p_c/(1-p_c)$, while the y-axis is the harmful/helpful amplitude ratio. A value below one means helpful updates are larger on average in that bin; a value above one means destructive updates are larger. The fitted positive slopes show that the ratio rises systematically with prior odds. This is why Table~\ref{tab:acs_params} reports positive $\alpha$ values: the destructive side grows faster than the constructive side as the model becomes more confident before tool use.

Figure~\ref{fig:appendix-delta-cost} then combines frequency and amplitude into the quantity most relevant for the main claim: the net expected update. In the left column, the positive component is the helpful mass and the negative component is the harmful mass; the black curve is their sum. The empirical zero crossing marks the point at which tool use stops increasing gold probability on average and starts decreasing it. In the right column, the same balance is expressed as a cost/benefit ratio. The horizontal zero line is the breakeven point on the $\log_{10}$ scale, so bars above zero indicate bins where harmful update mass exceeds helpful update mass.

Together, the two figures explain why average tool benefit can coexist with high-prior fragility. Tools repair many weak states, so the low-prior region can still be strongly positive. But when a strong correct belief is revised downward, the destructive move is large, and the local cost/benefit ratio can exceed one by orders of magnitude. This visual pattern is descriptive rather than a separate theorem; the threshold claim in the main text is supported by the sign-stable proxy splits in Tables~\ref{tab:boundary_transition_full}--\ref{tab:threshold-cross-model-additional}. The figures make the same mechanism easier to inspect: they show where the mass of revision changes from rescue-dominant to harm-dominant.

\FloatBarrier

\section{Operational Analyses for \textsc{CASE}}
\label{app:case}

This section retains the operational checks needed to support Section~\ref{sec:implication}: first-pair cost and disagreement audits, budget-ladder behavior, simple gate baselines, nearby answer-state signals, and the open-ended exact-match check. Less central token-boundary confidence tables were omitted from the main appendix to avoid a float backlog.

\subsection{Cost and first-disagreement audit}
\label{sec:case_cost_appendix}

\begin{table}[H]
\centering
\caption{\textbf{AT and NT sample cost audit.} Token proxy is stored serialized trajectory character count.}
\label{tab:case_cost_audit}
\footnotesize
\setlength{\tabcolsep}{3pt}
\resizebox{\linewidth}{!}{%
\begin{tabular}{lrrrr}
\toprule
\textbf{Side} & \textbf{Samples} & \textbf{Mean token proxy} & \textbf{Mean tool calls} & \textbf{Missing signal} \\
\midrule
AT & 141,956 & 1805.9 & 1.40 & 0.0\% \\
NT & 141,467 & 985.0 & 0.00 & 0.0\% \\
\bottomrule
\end{tabular}%
}
\end{table}

\begin{table}[H]
\centering
\caption{\textbf{First-pair disagreement audit.} Accuracy is measured only where first AT and first NT answers disagree; Mix-MV is random-tie mixed-majority expectation.}
\label{tab:case_disagreement_audit}
\scriptsize
\setlength{\tabcolsep}{2.5pt}
\resizebox{\linewidth}{!}{%
\begin{tabular}{lrrrrrrr}
\toprule
\textbf{Model} & \textbf{$N$} & \textbf{CASE} & \textbf{CISC} & \textbf{Mix-MV} & \textbf{Protect} & \textbf{Rescue} & \textbf{Both wrong} \\
\midrule
30B-NV & 3,897 & 39.2 & 29.2 & 27.7 & 71.2 & 70.5 & 44.7\% \\
Qwen3-Coder-30B & 6,975 & 37.7 & 27.5 & 28.2 & 81.3 & 51.1 & 43.7\% \\
Qwen3.5-9B & 6,905 & 34.0 & 20.9 & 28.9 & 60.4 & 57.4 & 42.2\% \\
\bottomrule
\end{tabular}%
}
\end{table}

The both-wrong column explains why disagreement accuracy is much lower than full-pool aggregate accuracy: in roughly half of first-pair disagreements, neither channel supplies the correct answer.
\FloatBarrier
\begin{table}[H]
\centering
\caption{\textbf{Stratified disagreement summaries supporting Section~\ref{sec:case_disagreement_main}.} Entries are unweighted averages across model settings on the first-pair AT/NT disagreement set. Protect is measured on NT-correct/AT-wrong cases; rescue is measured on NT-wrong/AT-correct cases.}
\label{tab:case_strata_support}
\footnotesize
\setlength{\tabcolsep}{4pt}
\begin{tabular}{lrrr}
\toprule
\textbf{Stratum} & \textbf{CASE-CISC gap} & \textbf{Protect} & \textbf{Rescue} \\
\midrule
$\eta\le\hat\nu$ & +7.7 & -- & -- \\
$\eta>\hat\nu$ & +13.5 & -- & -- \\
$p_{c,\mathrm{pre}}\ge0.80$ & +20.8 & 86.5 & 72.5 \\
\bottomrule
\end{tabular}
\end{table}

\subsection{Budget and fixed-budget behavior}

\begin{table*}[!t]
\centering
\caption{\textbf{Cross-model operating points.} Entries with ``@'' report accuracy at normalized cost.}
\label{tab:matched_cost}
\footnotesize
\setlength{\tabcolsep}{6pt}
\begin{tabular}{lrrrrrr}
\toprule
\textbf{Model} & \textbf{1+1 MV} & \textbf{1+1 Signal} & \textbf{8+4 MV} & \textbf{8+4 Signal} & \textbf{Best CASE $\leq1$} & \textbf{Best Seq $\leq1$} \\
\midrule
Qwen3-Coder-30B & 53.6 & 57.3 & 60.2 & 61.3 & 60.8 @ 0.749 & 60.8 @ 0.881 \\
30B-NV & 66.0 & 68.5 & 70.5 & 72.0 & 71.2 @ 0.806 & 70.3 @ 0.634 \\
Qwen3.5-9B & 44.0 & 46.0 & 47.6 & 49.6 & 48.7 @ 0.791 & 48.4 @ 0.912 \\
\bottomrule
\end{tabular}
\end{table*}

\begin{table}[H]
\centering
\caption{\textbf{Fixed-budget gains over mixed majority.} Each cell is Mixed Signal minus Mixed MV in aggregate accuracy points; QC abbreviates Qwen3-Coder-30B.}
\label{tab:fixed_budget}
\footnotesize
\setlength{\tabcolsep}{3.4pt}
\begin{tabular}{lrrrrrr}
\toprule
\textbf{Model} & \textbf{1+1} & \textbf{2+2} & \textbf{4+2} & \textbf{4+4} & \textbf{8+4} & \textbf{8+8} \\
\midrule
QC-30B & +3.8 & +1.9 & +1.6 & +1.6 & +1.1 & +1.6 \\
30B-NV & +2.5 & +1.4 & +1.5 & +1.4 & +1.5 & +1.4 \\
Q3.5-9B & +2.0 & +1.4 & +1.9 & +1.5 & +1.9 & +1.8 \\
\bottomrule
\end{tabular}
\end{table}

\begin{table}[H]
\centering
\caption{\textbf{P(True)-CISC fixed-budget deltas.} Each cell is P(True)-CISC minus mixed majority on the same candidate-source files; QC abbreviates Qwen3-Coder-30B.}
\label{tab:ptrue_cisc_budget}
\footnotesize
\setlength{\tabcolsep}{3.4pt}
\begin{tabular}{lrrrrrr}
\toprule
\textbf{Model} & \textbf{1+1} & \textbf{2+2} & \textbf{2AT} & \textbf{2NT} & \textbf{4+4} & \textbf{8+8} \\
\midrule
QC-30B & -0.3 & -0.2 & +0.1 & -0.3 & -0.2 & -0.1 \\
30B-NV & +0.6 & +0.2 & +0.6 & +0.4 & -0.1 & +0.0 \\
Q3.5-9B & -3.2 & -1.6 & -3.1 & -2.9 & -0.8 & -0.5 \\
\bottomrule
\end{tabular}
\end{table}

\subsection{Gate and signal ablations}
\label{sec:signal_choice_appendix}

\begin{table}[H]
\centering
\caption{\textbf{Simple prior-protection gates are insufficient.} Each entry is the best aggregate-accuracy point from a threshold sweep on the fixed Qwen3-Coder-30B export.}
\label{tab:gate_baselines}
\footnotesize
\setlength{\tabcolsep}{4pt}
\begin{tabular}{lccc}
\toprule
\textbf{Budget} & \textbf{Mix-cert} & \textbf{Freeze} & \textbf{Skip} \\
\midrule
1+1 & 57.3 @ 0.195 & 55.3 @ 0.157 & 54.7 @ 0.141 \\
2+2 & 59.5 @ 0.391 & 57.1 @ 0.303 & 56.6 @ 0.274 \\
\bottomrule
\end{tabular}
\end{table}

\begin{table}[H]
\centering
\caption{\textbf{Nearby answer-state signals nearly tie.} Accuracy is aggregate accuracy on the shared four-benchmark Qwen3-Coder-30B evaluation.}
\label{tab:signal_choice}
\footnotesize
\setlength{\tabcolsep}{4pt}
\begin{tabular}{lrr}
\toprule
\textbf{Signal} & \textbf{Acc.} & \textbf{vs MV} \\
\midrule
Normalized entropy certainty & 57.2 & +4.2 \\
softplus$(\mathrm{KL}(p_i\|p_f))$ & 57.3 & +4.2 \\
max$(c_{\mathrm{pre}}, c_{\mathrm{post}})$ & 57.3 & +4.2 \\
Post-probe margin & 57.2 & +4.2 \\
\bottomrule
\end{tabular}
\end{table}

\subsection{Free-form exact-match check}
\label{sec:freeform_stress}

The main mechanism analysis is identified on multiple-choice data, where gold-event probabilities and answer-space entropy are directly observable. As an operational check beyond closed-choice grading, we also evaluate AT/NT arbitration on boxed exact-match math runs.

\begin{table*}[!t]
\centering
\caption{\textbf{Free-form exact-match AT/NT arbitration.} Rows report matched-budget AUC over the $1{+}1,2{+}2,4{+}4,8{+}8$ ladder. The probe denominator is the retained completed-probe count; ten expected probe slots were absent after completion/parsing filters.}
\label{tab:freeform_stress}
\footnotesize
\setlength{\tabcolsep}{6pt}
\begin{tabular}{lrrrrrrr}
\toprule
\textbf{Dataset} & \textbf{$N$} & \textbf{Probe} & \textbf{Mix-MV} & \textbf{CASE} & \textbf{CISC} & \textbf{CASE-CISC} & \textbf{CISC-MV} \\
\midrule
AMC23 & 40 & 639/639 & 77.0 & 77.5 & 76.9 & +0.6 & -0.2 \\
AIME 2026 & 30 & 480/480 & 35.2 & 43.3 & 37.5 & +5.8 & +2.3 \\
BeyondAIME & 100 & 1591/1591 & 13.0 & 13.3 & 12.8 & +0.5 & -0.2 \\
\midrule
Macro & 170 & 2710/2710 & 41.7 & 44.7 & 42.4 & +2.3 & +0.7 \\
Pooled & 170 & 2710/2710 & 32.0 & 33.7 & 32.2 & +1.5 & +0.3 \\
\bottomrule
\end{tabular}
\end{table*}

\FloatBarrier
\section{Proofs and Additional Derivations}
\label{app:proofs}

This appendix records the proofs deferred from Section~\ref{sec:theory}. The derivations proceed from one-step logarithmic pooling to the threshold boundary condition and then to the finite-step destruction result used to interpret confident wrong endpoints.

\subsection{Minimal-change objective and logarithmic pool}
\label{app:log_pool_proof}

We adopt the same minimal-change principle used in Section~\ref{sec:theory}. The revised state should remain close to both the internal belief and the contextual belief, with $\lambda_x$ determining how far revision is allowed to move toward the external channel:
\begin{equation}
\begin{aligned}
R_x^{\lambda}
=\arg\min_{R \in \Delta(\mathcal{A})}\Big[&
(1-\lambda_x)D_{\mathrm{KL}}(R\,\|\,P_x)\\
&+\lambda_xD_{\mathrm{KL}}(R\,\|\,Q_x)\Big].
\end{aligned}
\label{eq:revobj}
\end{equation}

\begin{theorem}[Minimal-change revision yields a logarithmic pool]
\label{thm:logpool}
For any finite answer space $\mathcal{A}$, distributions $P_x,Q_x$ with common positive support, and any $\lambda_x \in [0,1]$, the optimization problem in Eq.~\eqref{eq:revobj} has a unique solution
\begin{equation}
R_x^{\lambda}(a)
=
\frac{P_x(a)^{1-\lambda_x}Q_x(a)^{\lambda_x}}
{\sum_{b\in\mathcal{A}}P_x(b)^{1-\lambda_x}Q_x(b)^{\lambda_x}},
\qquad a\in\mathcal{A}.
\label{eq:log_pool}
\end{equation}
That is, the revised belief is a logarithmic opinion pool.
\end{theorem}

The objective in Eq.~\eqref{eq:revobj} is strictly convex in $R$ over the simplex because the KL divergence is strictly convex in its first argument and the weights are nonnegative. Introduce the Lagrangian
\begin{equation}
\begin{aligned}
\mathcal{J}(R,\tau)
&=(1-\lambda)D_{\mathrm{KL}}(R\|P)
+\lambda D_{\mathrm{KL}}(R\|Q) \\
&\quad +\tau\left(\sum_{a\in\mathcal{A}}R(a)-1\right).
\end{aligned}
\end{equation}
For any answer $a$ with positive support under both $P$ and $Q$, stationarity gives
\begin{equation}
\begin{aligned}
&(1-\lambda)\left(\log\frac{R(a)}{P(a)}+1\right)\\
&\quad+\lambda\left(\log\frac{R(a)}{Q(a)}+1\right)+\tau=0.
\end{aligned}
\end{equation}
Collecting terms, exponentiating, and normalizing over $a$ yields Eq.~\eqref{eq:log_pool}.

\subsection{Binary log-odds form}

Collapse a multiple-choice question to the binary event ``gold answer $y^\star$'' versus ``non-gold.'' Let
\begin{equation}
p=P_x(y^\star), \qquad q=Q_x(y^\star), \qquad r=R_x^{\lambda}(y^\star),
\end{equation}
and define log-odds $\eta=\operatorname{logit}(p)$, $\nu=\operatorname{logit}(q)$, and $\rho=\operatorname{logit}(r)$.

\begin{corollary}[Gold-event log-odds update]
\label{cor:log_odds}
Under the binary gold-vs.-rest reduction,
\begin{equation}
\rho = (1-\lambda)\eta + \lambda \nu,
\label{eq:logodds_update}
\end{equation}
and therefore
\begin{equation}
\rho - \eta = \lambda(\nu-\eta).
\label{eq:logodds_shift}
\end{equation}
\end{corollary}

The derivation follows by applying the logarithmic pool to the binary odds ratio:
\begin{equation}
\frac{R^{\lambda}(c)}{R^{\lambda}(\neg c)}
=
\left(\frac{p}{1-p}\right)^{1-\lambda}
\left(\frac{q}{1-q}\right)^\lambda .
\end{equation}
Taking logs yields Eq.~\eqref{eq:logodds_update}, and Eq.~\eqref{eq:logodds_shift} follows by subtracting $\eta$.

\subsection{Proof of Theorem~\ref{thm:threshold}}

Let
\begin{equation}
Z_{\lambda}=(1-\lambda)\eta+\lambda M,
\qquad
\ell(z)=\log(1+e^{-z}).
\end{equation}
Then the gold-event logistic risk is
\begin{equation}
\mathcal{L}(\lambda;\eta)
=
\mathbb{E}_{M}\!\left[\log\!\left(1+\exp(-Z_{\lambda})\right)\right].
\label{eq:risk}
\end{equation}
Because $\ell'(z)=-\sigma(-z)$ and $\partial Z_{\lambda}/\partial \lambda=M-\eta$,
\begin{equation}
\frac{\partial \mathcal{L}}{\partial \lambda}
=
-\mathbb{E}\left[\sigma(-Z_{\lambda})(M-\eta)\right].
\end{equation}
Differentiating again gives
\begin{equation}
\frac{\partial^2 \mathcal{L}}{\partial \lambda^2}
=
\mathbb{E}\!\left[\sigma(Z_{\lambda})\sigma(-Z_{\lambda})(M-\eta)^2\right] \ge 0,
\end{equation}
so $\mathcal{L}$ is convex in $\lambda$. Evaluating at $\lambda=0$ gives
\begin{equation}
\begin{aligned}
\left.\frac{\partial \mathcal{L}}{\partial \lambda}\right|_{\lambda=0}
&= -\sigma(-\eta)\,\mathbb{E}[M-\eta]\\
&=-(1-p)(\mathbb{E}[M]-\eta).
\end{aligned}
\end{equation}
If $\eta\ge\mathbb{E}[M]$, the derivative at the boundary is nonnegative and convexity implies that $\lambda=0$ is globally optimal. If $\eta<\mathbb{E}[M]$, the boundary derivative is negative, so sufficiently small positive $\lambda$ lowers risk.

\subsection{Iterated destruction under an authority floor}
\label{app:proofs:iterated}

The main text keeps the theory focused on the threshold law, but the wrong-trace audit raises a follow-up question: why can harmful trajectories stay confident while becoming wrong? A simple iterative extension answers that question:
\begin{equation}
\eta_{t+1}=(1-\lambda_t)\eta_t+\lambda_t \nu_t.
\label{eq:iterative_update}
\end{equation}
Here $\eta_t$ is the gold log-odds before revision step $t$, $\nu_t$ is the effective contextual log-odds introduced at that step, and $\lambda_t$ is the authority granted to it.

\begin{theorem}[Finite-step belief destruction under an authority floor]
\label{thm:finite_step}
Suppose $\eta_0>0$ is initially correct and strong. If there exist constants $\underline{\lambda}>0$ and $\bar{\nu}<0$ such that for all $t$,
\begin{equation}
\lambda_t \ge \underline{\lambda},
\qquad
\nu_t \le \bar{\nu},
\end{equation}
then
\begin{equation}
\eta_t
\le
(1-\underline{\lambda})^t\eta_0
+
\big(1-(1-\underline{\lambda})^t\big)\bar{\nu}.
\label{eq:finite_step_bound}
\end{equation}
In particular, after finitely many steps the gold log-odds becomes negative, so the correct belief flips to an incorrect one.
\end{theorem}

\subsection{Proof of Theorem~\ref{thm:finite_step}}

Define
\begin{equation}
B_t=(1-\underline{\lambda})^t\eta_0+
\bigl(1-(1-\underline{\lambda})^t\bigr)\bar{\nu}.
\label{eq:iterated_bound}
\end{equation}
We prove $\eta_t\le B_t$ by induction. The base case is immediate because $B_0=\eta_0$. Moreover, $B_t\ge\bar{\nu}$ for every $t$, since it is a convex combination of $\eta_0$ and $\bar{\nu}$. If $\eta_t\le B_t$, then Eq.~\eqref{eq:iterative_update} and the assumptions $\nu_t\le\bar{\nu}$ and $\lambda_t\ge\underline{\lambda}$ give
\begin{align}
\eta_{t+1}
&\le (1-\lambda_t)B_t+\lambda_t\bar{\nu} \\
&\le (1-\underline{\lambda})B_t+\underline{\lambda}\bar{\nu}
 = B_{t+1},
\end{align}
where the second inequality uses $B_t\ge\bar{\nu}$. This establishes Eq.~\eqref{eq:finite_step_bound}. Because $B_t\to\bar{\nu}<0$, the upper bound becomes negative after finitely many steps, and hence so does $\eta_t$.

\subsection{Scope of the theoretical result}

The threshold law explains why sign reversal and high-prior fragility are expected under a misallocated revision policy, and why harmful traces can remain confident. The empirical power law in Eq.~\eqref{eq:destruction_law} remains an observed scaling relation rather than a derived theorem.

\FloatBarrier
\section{Experimental Details and Protocols}
\label{app:details}

This section records the retained reproducibility fields and the two protocols that are most directly tied to the main-text claims: the wrong-trace audit and PAIT. The original long prompt-template tables have been condensed into prose to avoid several consecutive full-width floats.

\subsection{Model identifiers and fixed configuration}

All three models are open-source Hugging Face checkpoints. We report the public
model identifiers and all retained run settings to make the experimental protocol
rerunnable. All runs were served with SGLang and executed on two NVIDIA A800 GPUs. Because the experiments use public Hugging Face checkpoints, a fixed reported
protocol, and a consistent SGLang serving setup, the runs
are reproducible at the protocol level. Throughout the paper, \emph{Physics} denotes our multiple-choice reformulation of the HiPhO high-school physics Olympiad benchmark \citep{yu2025hipho}.

\begin{table}[H]
\centering
\caption{\textbf{Fixed experimental configuration.} Defaults are taken from launch scripts unless overridden by the recorded run configuration.}
\label{tab:repro}
\footnotesize
\setlength{\tabcolsep}{3pt}
\renewcommand{\arraystretch}{1.10}
\begin{tabularx}{\linewidth}{L{0.30\linewidth}Y}
\toprule
\textbf{Field} & \textbf{Value} \\
\midrule
Main mechanism run & Qwen3-Coder-30B on MMLU-Pro, GPQA Diamond, Physics, and SuperGPQA-Calc. \\
Cross-model runs & 30B-NV\citep{nvidia2025nemotron3nano30b} and Qwen3.5-9B\citep{qwen2026qwen35} on the same benchmark set. \\
Free-form runs & AMC23, AIME 2026, and BeyondAIME exact-match math. \\
Settings & NT uses no tools; AT uses \texttt{always\_tool} without SSGR rollback. \\
AT / NT samples & 8 / 8 per question; \texttt{n\_rollouts=1}. \\
Sampling & \texttt{temperature=0.7}; \texttt{seed=42}; probes and finalization are greedy. \\
Context / timeout & \texttt{max\_model\_len=81920}; per-request timeout \texttt{1800s}; max retries \texttt{8}. \\
Tool-call budget & At most \texttt{5} tool calls per trajectory in the main batch launch. \\
Tools & \texttt{run\_python}; \texttt{wiki\_search} when the Wiki service is enabled. \\
MCQ state probe & Guided option choice; \texttt{max\_tokens=1}; \texttt{temperature=0}; thinking disabled. \\
Main support proxy & Leave-one-out mean post-tool $p_c$ within matched event strata. \\
\bottomrule
\end{tabularx}
\end{table}

\subsection{Tool execution parameters}
\label{sec:tool_params}

\begin{table}[H]
\centering
\caption{\textbf{Tool execution parameters.} The Wiki service retrieves pre-indexed corpus rows/chunks; \texttt{max\_chars} only truncates returned snippets.}
\label{tab:tool_params}
\footnotesize
\setlength{\tabcolsep}{3pt}
\renewcommand{\arraystretch}{1.10}
\begin{tabularx}{\linewidth}{L{0.24\linewidth}L{0.25\linewidth}Y}
\toprule
\textbf{Tool} & \textbf{Parameter} & \textbf{Value / behavior} \\
\midrule
\texttt{run\_python} & Backend & Local subprocess executor; no Docker container. \\
\texttt{run\_python} & Timeout & Executor \texttt{30s}, wrapped by async \texttt{60s} guard. \\
\texttt{run\_python} & Limits / prelude & 16GB memory limit; common math/science libraries auto-imported if installed. \\
\texttt{run\_python} & Observation & JSON stdout/stderr/exit/time; long fields truncated by runner packaging. \\
\texttt{wiki\_search} & Backend & Local FastAPI service backed by E5 embeddings and FAISS. \\
\texttt{wiki\_search} & \texttt{topk} & Model-generated values normalized to \texttt{1..5}; runner default \texttt{3}. \\
\texttt{wiki\_search} & \texttt{max\_chars} & Model values clipped to \texttt{50..100000}; runner default \texttt{12000}. \\
\texttt{wiki\_search} & Errors & Retries configured Wiki URLs up to three times, then returns empty-result error JSON. \\
\bottomrule
\end{tabularx}
\end{table}

\subsection{Wrong-trace audit protocol}
\label{sec:silver_label_annotation}

Section~\ref{sec:silver} uses two wrong-trace audits. The first is an automatic surface-plausibility audit over the four headline datasets. It labels each AT trajectory by the apparent relation between the tool evidence and the model's final answer: supportive, contradictory, irrelevant, or other. The labeller is instructed to use only the provided fields (question, answer options, model final answer, assistant text before the tool, tool calls and outputs, and assistant text after the tool). It returns a single JSON object with evidence relation, tool-use quality, failure mode, confidence, spans, and a short rationale.

The automatic audit covers 13,390 labelled AT traces, of which 5,725 end in a wrong final answer. Among these wrong AT traces, 2,098/5,725 (36.6\%) are supportive-looking, 1,964/5,725 (34.3\%) are contradictory, 1,550/5,725 (27.1\%) are irrelevant, and 113/5,725 (2.0\%) are other. Independently of the automatic evidence-status label, we define ``high-confidence'' using the stored final answer-space probe: the probability assigned to the trajectory's own wrong answer is at least $\tau_{\mathrm{conf}}=0.8$. Among wrong traces whose evidence is contradictory or irrelevant, 3,452/3,514 (98.2\%) remain high-confidence by this probe-based criterion.

\begin{table}[H]
\centering
\caption{\textbf{Automatic surface-plausibility audit over wrong AT traces.} The left block reports evidence-status shares among wrong AT traces. The right block reports probe-based high-confidence rates among non-supportive wrong traces.}
\label{tab:silver_surface_audit_appendix}
\footnotesize
\setlength{\tabcolsep}{2pt}
\begin{tabularx}{\linewidth}{L{0.47\linewidth}rrr}
\toprule
\textbf{Quantity} & \textbf{Num.} & \textbf{Denom.} & \textbf{Rate} \\
\midrule
Supportive-looking wrong traces & 2,098 & 5,725 & 36.6\% \\
Contradictory wrong traces & 1,964 & 5,725 & 34.3\% \\
Irrelevant wrong traces & 1,550 & 5,725 & 27.1\% \\
Other wrong traces & 113 & 5,725 & 2.0\% \\
\midrule
Contradictory wrong traces that remain high-confidence & 1,945 & 1,964 & 99.0\% \\
Irrelevant wrong traces that remain high-confidence & 1,507 & 1,550 & 97.2\% \\
Contradictory/irrelevant combined & 3,452 & 3,514 & 98.2\% \\
\bottomrule
\end{tabularx}
\end{table}

The second audit is the retained mechanism-source audit. Its sampling frame is wrong AT trajectories, not all harmful trajectories. We first filter to AT samples whose final answer is wrong, that contain at least one non-empty and non-error tool call, and whose final assistant response is substantive. We then draw 240 traces from three model sources, with 80 traces per model (Qwen3-Coder-30B, 30B-NV, and Qwen3.5-9B). Sampling is stratified by an automatic pre-screen over truth-axis labels, used only to improve coverage of candidate mechanisms; the final mechanism labels are assigned independently. 

For each retained trace, the annotator sees the question, answer options, gold answer, model's wrong final answer, each tool output, the trace window immediately following the tool call, and the final assistant message. The five source labels are: A, tool-correct but answer-space reinterpreted; B, tool-partial and taken at face value; C, tool-irrelevant but confabulated into support; D, tool-contradictory but absorbed anyway; and U, unclear. The second binary label records whether the final wrong answer remains a high-confidence wrong state.

To estimate annotation stability, three experts audit the same 240-item manifest independently. For the five-way mechanism label, Fleiss' $\kappa$ across the three same-model rounds is 0.485, with pairwise Cohen's $\kappa$ values of 0.507, 0.467, and 0.489. For the binary high-confidence-wrong label, Fleiss' $\kappa$ is 0.617, with pairwise Cohen's $\kappa$ values of 0.627, 0.591, and 0.637. We report the majority-vote labels below.

\begin{table}[H]
\centering
\caption{\textbf{Mechanism-source distribution in the 240-item audit.} Percentages are within model; the overall row pools all sampled items.}
\label{tab:silver_type_distribution_appendix}
\scriptsize
\setlength{\tabcolsep}{3pt}
\resizebox{\linewidth}{!}{%
\begin{tabular}{lrrrrrr}
\toprule
\textbf{Model} & \textbf{$n$} & \textbf{A} & \textbf{B} & \textbf{C} & \textbf{D} & \textbf{U} \\
\midrule
30B-NV & 80 & 3.8\% & 16.2\% & 36.2\% & 16.2\% & 27.5\% \\
Qwen3-Coder-30B & 80 & 25.0\% & 17.5\% & 2.5\% & 55.0\% & 0.0\% \\
Qwen3.5-9B & 80 & 16.2\% & 12.5\% & 35.0\% & 32.5\% & 3.8\% \\
\midrule
Overall & 240 & 15.0\% & 15.4\% & 24.6\% & 34.6\% & 10.4\% \\
\bottomrule
\end{tabular}%
}
\end{table}

\begin{table}[H]
\centering
\caption{\textbf{High-confidence-wrong rates and eta-regime counts.} HC wrong is the Step-2 majority label; the eta-regime split uses the main $\eta>\hat\nu$ proxy.}
\label{tab:silver_hc_eta_appendix}
\scriptsize
\setlength{\tabcolsep}{3pt}
\begin{tabular}{lrrrrr}
\toprule
\textbf{Quantity} & \textbf{A} & \textbf{B} & \textbf{C} & \textbf{D} & \textbf{U} \\
\midrule
HC wrong rate & 55.6\% & 29.7\% & 41.1\% & 57.8\% & 20.0\% \\
$\eta>\hat\nu$ count & 15 & 10 & 14 & 29 & 5 \\
$\eta\le\hat\nu$ count & 21 & 27 & 45 & 54 & 20 \\
\bottomrule
\end{tabular}
\end{table}

Overall, 45.4\% of sampled wrong AT traces are labelled high-confidence wrong, with model-level rates of 25.0\% for 30B-NV, 47.5\% for Qwen3-Coder-30B, and 63.8\% for Qwen3.5-9B. The mechanism distribution differs strongly by model ($\chi^2=88.378$, $df=8$, $p=1.0\times10^{-15}$).

\section{Prompt Template For Tool Augmented Reasoning}
\label{app:prompt-template}

\begin{promptbox}{System Prompt for Tool-Augmented Problem Solving}
You are an expert problem-solving agent with access to tools via function calling.
Your task is to solve questions across domains including science, engineering,
medicine, economics, law, history, mathematics, and related fields.

Reasoning protocol:
Work iteratively using a ReAct-style procedure:
(1) Think: analyze the problem and identify what information or computation is needed.
(2) Call Tool: use the appropriate tool when verification or computation is required.
(3) Observe: inspect the returned result carefully.
(4) Repeat or Answer: continue using tools until confident, then provide the final answer.

Key principle:
Do not guess. Use tools to verify factual claims, calculations, and intermediate results.

Available tools:

1. run_python
Purpose:
Execute Python code in a sandboxed environment.

Input schema:
{"code": "<python code as string>"}

Output schema:
{"stdout": "...", "stderr": "...", "exit_code": 0, "exec_time": 0.xx}

Use run_python for:
- multi-step arithmetic or algebraic computation;
- numerical comparison across answer options;
- unit conversion or dimensional analysis;
- probability, statistics, or combinatorics;
- data processing, parsing, or verification;
- symbolic or formula-based calculation.

2. wiki_search
Purpose:
Search a local Wikipedia knowledge base for factual information.

Input schema:
{"query": "<search query>", "topk": 1, "max_chars": 12000}

Output schema:
{"results": [{"title": "...", "content": "...", "score": 0.xx}, ...]}

Use wiki_search for:
- dates, names, events, definitions, or named entities;
- historical, scientific, medical, or legal facts;
- biographies, mechanisms, laws, or theoretical concepts;
- any factual claim about which you are uncertain.

Tool-use policy:
- Always use run_python for calculations requiring more than one arithmetic or algebraic step.
- Always use wiki_search for specific factual verification.
- For complex questions requiring both factual context and computation, first use wiki_search
  to gather the needed facts, then use run_python to compute or verify the answer.
- If a tool call fails, inspect the error and retry with a corrected call when possible.

Tool-call format:
Use the following XML-style wrapper with JSON content.

<tool_call>
{"name": "run_python", "arguments": {"code": "<python code here>"}}
</tool_call>

<tool_call>
{"name": "wiki_search", "arguments": {"query": "<search query>", "topk": 1}}
</tool_call>

Final answer format:
For multiple-choice questions, output the final answer as:
\boxed{X}

where X is the single letter corresponding to the correct option.

For open-ended numerical questions, output the final result in boxed form:
\boxed{<answer>}

Important reminders:
- Use tools proactively to reduce reasoning errors.
- Keep tool calls targeted and relevant.
- Verify intermediate results before giving the final answer.
- Stay focused on the question and avoid unnecessary elaboration.
\end{promptbox}

\newpage
\subsection{PAIT protocol and selection flow}
\label{sec:pait_protocol}

PAIT is a same-content, different-wrapper intervention. For each selected high-prior harmful case, we identify the misleading information in the stored AT trajectory that supports the historical wrong answer and rewrite it into a short, wrapper-invariant clue, stored as \texttt{adv\_claim}. The rewrite preserves the substantive content and target answer of the original trajectory while removing trajectory-specific wording so that the same clue can be reused across conditions. The resulting \texttt{adv\_claim} is inserted verbatim into the \textsc{Tool}, \textsc{Context}, \textsc{Reflection}, and \textsc{Tool+Antidote} answer prompts.

For the human annotation version, annotators are shown the dataset and item identifier, question, options, option letters, gold answer, historical harmful final answer, pre/post gold-answer probabilities, prior bin, and the harmful trajectory transcript. They do not change the question, options, or gold label. Their task is to identify the misleading information associated with the harmful revision and rewrite it as one short, plausible clue that conflicts with the gold answer, points to a single non-gold option, and can be reused word-for-word under every wrapper.

\begin{table*}[tbp]
\centering
\caption{\textbf{Human PAIT adversarial-clue annotation form.} These are the fields recorded for each selected high-prior harmful example.}
\label{tab:pait_human_form}
\footnotesize
\setlength{\tabcolsep}{4pt}
\begin{tabularx}{\textwidth}{L{0.20\textwidth}L{0.18\textwidth}Y}
\toprule
\textbf{Field} & \textbf{Type} & \textbf{Acceptance rule} \\
\midrule
\texttt{adv\_claim} & Free text & One or two sentences, preferably under 40 words; rewritten from the misleading information in the harmful AT trajectory; false relative to the gold answer; no mention of tools, Python, search, prompts, logs, or the model. \\
\texttt{target\_wrong\_letter} & Single choice & Must be exactly one non-gold option, preferably the historical harmful final answer when that direction is clear. \\
\texttt{claim\_type} & Single choice & One of \texttt{numeric}, \texttt{comparison}, \texttt{definition}, \texttt{causal}, or \texttt{other}. \\
\texttt{why\_plausible} & Free text & Briefly explains why the clue sounds like a valid calculation, comparison, definition, or causal fact. \\
\texttt{source\_support} & Free text & Briefly identifies the harmful-trace mistake or tool-output interpretation from which the clue was rewritten. \\
\bottomrule
\end{tabularx}
\end{table*}

\begin{table*}[tbp]
\centering
\caption{\textbf{PAIT wrapper templates.} The same \texttt{ADV\_CLAIM} string is inserted without rewriting. All probes use the same one-letter option-choice format.}
\label{tab:pait_wrapper_templates}
\footnotesize
\setlength{\tabcolsep}{4pt}
\begin{tabularx}{\textwidth}{L{0.17\textwidth}L{0.27\textwidth}Y}
\toprule
\textbf{Condition} & \textbf{Wrapper cue} & \textbf{Controlled contrast} \\
\midrule
Baseline & No external signal; solve from the question and options. & Estimates the clean unrevised answer state. \\
Tool & ``An external tool now reports: \texttt{ADV\_CLAIM}.'' & Presents the clue with external-tool authority. \\
Context & ``Additional reference information: \texttt{ADV\_CLAIM}.'' & Presents the identical clue as reference context. \\
Reflection & ``Your earlier reasoning may have overlooked the following clue: \texttt{ADV\_CLAIM}.'' & Presents the identical clue as a self-correction cue. \\
Tool+Antidote & Tool wrapper plus a warning that external tools can hallucinate or return incorrect numerical results. & Tests whether explicitly limiting tool authority attenuates the same-clue damage. \\
\bottomrule
\end{tabularx}
\end{table*}

\FloatBarrier

\begin{table}[H]
\centering
\caption{\textbf{PAIT selection flow.} Selection starts from the Qwen3-Coder-30B live run and then deduplicates by question before balanced per-dataset sampling.}
\label{tab:pait_selection}
\scriptsize
\setlength{\tabcolsep}{3pt}
\begin{tabular}{lrrrr}
\toprule
\textbf{Dataset} & \textbf{Seen} & \textbf{Cand.} & \textbf{Deduped} & \textbf{Selected} \\
\midrule
MMLU-Pro & 6,696 & 922 & 258 & 149 \\
GPQA Diamond & 198 & 109 & 29 & 29 \\
Physics & 325 & 102 & 31 & 31 \\
SuperGPQA-Calc & 11,229 & 1,206 & 282 & 166 \\
\midrule
Total & 18,448 & 2,339 & 600 & 375 \\
\bottomrule
\end{tabular}
\end{table}

\begin{table}[H]
\centering
\caption{\textbf{Detailed cross-model PAIT on eligible high-prior cases.} FR is answer flip rate under the adversarial-clue wrapper; parentheses report eligible rows.}
\label{tab:pait-cross-model-detail}
\scriptsize
\setlength{\tabcolsep}{2.5pt}
\resizebox{\linewidth}{!}{%
\begin{tabular}{lcccc}
\toprule
\textbf{Model} & \textbf{Tool FR} & \textbf{Context FR} & \textbf{Reflection FR} & \textbf{Tool+Antidote FR} \\
\midrule
Qwen3-Coder-30B & 77.0\% (104) & 63.7\% (104) & 78.6\% (104) & 37.4\% (104) \\
Qwen3.5-9B & 20.5\% (163) & 41.9\% (163) & 85.0\% (163) & 0.0\% (163) \\
30B-NV & 51.9\% (259) & 26.8\% (259) & 44.1\% (260) & 20.0\% (250) \\
\bottomrule
\end{tabular}%
}
\end{table}

\begin{table}[H]
\centering
\caption{\textbf{Matched Qwen3-Coder-30B PAIT with no-claim control.} FR is answer flip rate on the condition-complete subset ($n=104$ per condition).}
\label{tab:pait-qwen-matched-detail}
\scriptsize
\setlength{\tabcolsep}{2.5pt}
\resizebox{\linewidth}{!}{%
\begin{tabular}{lrrrr}
\toprule
\textbf{Condition} & \textbf{Adv FR} & \textbf{Adv post $p_c$} & \textbf{No-claim FR} & \textbf{No-claim post $p_c$} \\
\midrule
Tool & 77.0\% & 0.253 & 2.0\% & 0.981 \\
Context & 63.7\% & 0.386 & 5.9\% & 0.942 \\
Reflection & 78.6\% & 0.251 & 9.8\% & 0.903 \\
Tool+Antidote & 37.4\% & 0.625 & 7.2\% & 0.942 \\
\bottomrule
\end{tabular}%
}
\end{table}
\FloatBarrier

The number of eligible PAIT rows differs across model settings because PAIT keeps only cases that pass the executed clean-baseline filter and the wrapper/probe completion filter. We therefore report flip rates with per-condition sample sizes rather than forcing a fixed $n$ across models.

\section{Additional Matched Audits}
\label{app:review_additions}

 The analyses below further test possible alternatives, quantify sampling uncertainty, validate an automatic audit.

\subsection{Confidence- and difficulty-matched crossing nulls}

The baseline null keeps each event's pre-tool logit fixed and shuffles observed logit updates within the original matching cells, preserving update magnitude while severing the event-level association. The stricter null shuffles within dataset-specific difficulty quartiles formed from each question's mean pre-tool gold probability. Crossing denotes $p_c>0.5\rightarrow p_c<0.5$.

\begin{table}[H]
\centering
\caption{\textbf{Matched null audit for harmful boundary crossings.} The reported $p$-value is the fraction of 1,000 null draws at or above the observed crossing rate.}
\label{tab:matched_null_appendix}
\scriptsize
\setlength{\tabcolsep}{2.2pt}
\resizebox{\linewidth}{!}{%
\begin{tabular}{llllrrr}
\toprule
\textbf{Slice} & \textbf{Shuffle within} & \textbf{Matched factor} & \textbf{Observed} & \textbf{Null mean} & \textbf{$p$} \\
\midrule
$\eta>\hat\nu$ ($N=13{,}756$) & dataset $\times$ tool-call index & update magnitude & 15.33\% & 6.09\% & $<.001$ \\
$\eta>\hat\nu$ ($N=13{,}756$) & dataset $\times$ difficulty & difficulty & 15.33\% & 10.19\% & $<.001$ \\
$p_{c,\mathrm{pre}}\ge0.80$ ($N=11{,}521$) & dataset $\times$ tool-call index & update magnitude & 16.32\% & 5.52\% & $<.001$ \\
$p_{c,\mathrm{pre}}\ge0.80$ ($N=11{,}521$) & dataset $\times$ difficulty & difficulty & 16.32\% & 10.32\% & $<.001$ \\
\bottomrule
\end{tabular}%
}
\end{table}

\subsection{Direct human validation of the automatic evidence-status audit}

A separate sample of 100 wrong AT traces was annotated by humans using the same supportive/contradictory/irrelevant/other taxonomy as the GPT-5 audit. This check assesses the automatic labels directly and is distinct from the 240-trace mechanism-source audit in Appendix~\ref{sec:silver_label_annotation}.

\begin{table}[H]
\centering
\caption{\textbf{Human validation of GPT-5 evidence-status labels.} Agreement is computed on the common 100-trace sample.}
\label{tab:human_label_validation_appendix}
\footnotesize
\begin{tabular}{lr}
\toprule
\textbf{Statistic} & \textbf{Value} \\
\midrule
Validation sample size & 100 \\
GPT-5 vs. human-majority exact agreement & 72\% \\
GPT-5 vs. human-majority Cohen's $\kappa$ & 0.51 \\
Human inter-annotator Fleiss' $\kappa$ & 0.83 \\
\bottomrule
\end{tabular}
\end{table}

\subsection{Aligned CASE--CISC uncertainty audit}

The aligned comparison pairs one AT and one NT candidate by model, dataset, and question identifier, removes incomplete pairs, and uses a model- and dataset-stratified paired bootstrap with 5,000 resamples. Both controllers are deterministic conditional on the retained candidate pools, so the intervals quantify question-sampling uncertainty.

\begin{table}[H]
\centering
\caption{\textbf{Pair-complete 1-vs.-1 CASE--P(True)-CISC comparison.}}
\label{tab:case_bootstrap_appendix}
\scriptsize
\setlength{\tabcolsep}{3pt}
\resizebox{\linewidth}{!}{%
\begin{tabular}{lrr}
\toprule
\textbf{Scope} & \textbf{CASE--CISC} & \textbf{95\% CI} \\
\midrule
Pooled, 53,184 matched model--question instances & +3.80 & [3.54, 4.05] \\
Qwen3-Coder-30B & +4.02 & [3.54, 4.49] \\
30B-NV & +2.20 & [1.86, 2.56] \\
Qwen3.5-9B & +5.18 & [4.67, 5.68] \\
Non-GPQA pooled & +3.84 & [3.58, 4.10] \\
GPQA pooled & 0.00 & [-2.37, 2.37] \\
\bottomrule
\end{tabular}%
}
\end{table}

The overall and model-level intervals exclude zero. The GPQA-specific interval includes zero, so its point estimate should be read as inconclusive. In the matched sample, each model contributes only 195--198 GPQA questions, and 67 of 210 first-pair disagreements are wrong in both channels, limiting recoverable headroom.

\end{document}